%% file: tro_sfahmi_19.tex
\documentclass[letterpaper, 10 pt, journal, twoside]{IEEEtran}
\input{includes/used_packages.tex}
\input{includes/acronyms.tex}

\input{includes/commands.tex}

\makeglossaries
\newcommand{\rev}[1]{\textcolor{black}{#1}}
\newcommand{\frev}[1]{\textcolor{black}{#1}}

\title{STANCE: Locomotion Adaptation over Soft Terrain}
\author{Shamel Fahmi, Michele Focchi, Andreea Radulescu, Geoff Fink, Victor Barasuol, and Claudio Semini
\thanks{
Manuscript received: April 26, 2019; 
Revised August 22, 2019;
Accepted October 2, 2019.
This paper was recommended for publication by 
Associate Editor X. XXXX and
Editor X. XXXX upon
evaluation of the  reviewers' comments.
(\textit{Corresponding author: Shamel Fahmi.})}
\thanks{All the authors are with the Dynamic Legged Systems Lab, Istituto Italiano di Tecnologia (IIT), Genova, 
	Italy (email: firstname.lastname@iit.it).}
}
\usepackage{pdfpages}
\begin{document}
\null
\includepdf[pages=-]{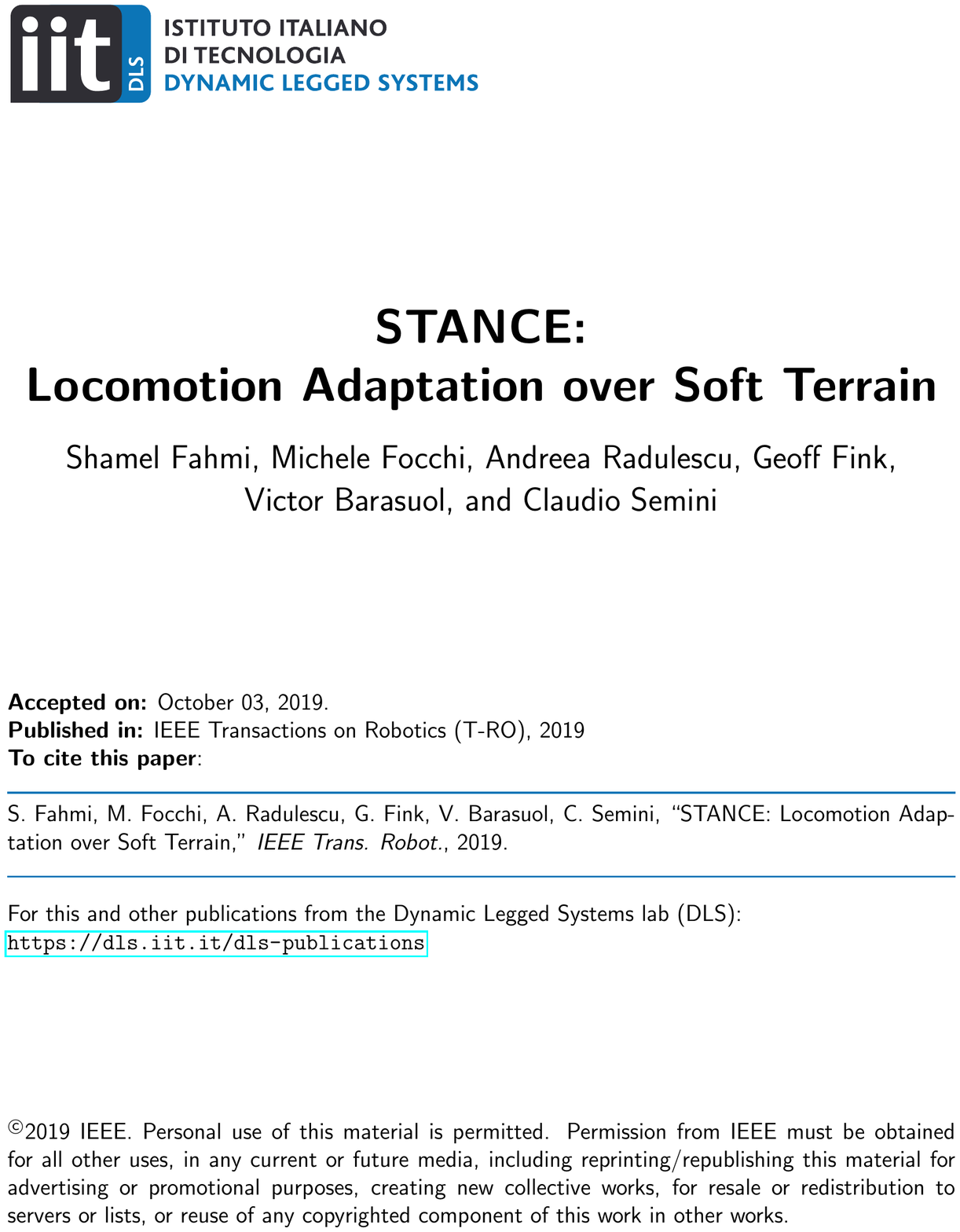}
\setcounter{page}{1}
\maketitle

\begin{abstract}
Whole-Body Control (WBC) has emerged as an important framework in locomotion control for legged robots.
However, most WBC frameworks fail to generalize beyond rigid terrains.  
Legged locomotion over soft terrain is difficult due to the presence of unmodeled contact dynamics 
that WBCs do not account for.
This introduces uncertainty in locomotion and affects the stability and performance of the system. 
In this paper, we propose a novel soft terrain adaptation algorithm called 
STANCE: Soft Terrain Adaptation and Compliance Estimation.
STANCE consists of a WBC that exploits the knowledge of the terrain to generate an optimal solution that is contact consistent
and an online terrain compliance estimator that provides the WBC with terrain knowledge.
We validated STANCE both in simulation and experiment on  the \gls{hyq} robot, and we compared it against the state of the art WBC. 
We demonstrated the capabilities of STANCE with multiple terrains of different compliances, aggressive maneuvers, different forward velocities, 
and external disturbances. STANCE allowed \gls{hyq} to adapt online to terrains with different compliances (rigid and soft) without pre-tuning. 
\gls{hyq} was able to  successfully deal with the transition between different terrains and showed the ability to  differentiate between compliances 
under each foot. 
\end{abstract}
\section{Introduction}\label{sec:introduction}
\IEEEPARstart{W}{hole-Body Control}
\gls{wbc} frameworks have achieved remarkable results in legged locomotion 
control~{\cite{Farshidian2017a,Bellicoso2017,Fahmi2019}}. 
Their main feature is that they use optimization techniques
to solve the locomotion control problem. 
\gls{wbc} can achieve multiple tasks in an optimal fashion 
by exploiting the robot's full dynamics and reasoning about both the actuation constraints and the contact 
interaction. 
These tasks include balancing,  interacting with the 
environment, and 
performing dynamic locomotion over a wide variety of terrains \cite{Fahmi2019}.
The tasks are executed at the robot's end effectors, 
but can also be utilized for contacts anywhere on the robot's body \cite{Henze2017} or for
a cooperative manipulation task between robots \cite{Bouyarmane2018}.

\begin{figure}[tb]
	\centering
	\includegraphics[width=0.485\textwidth]{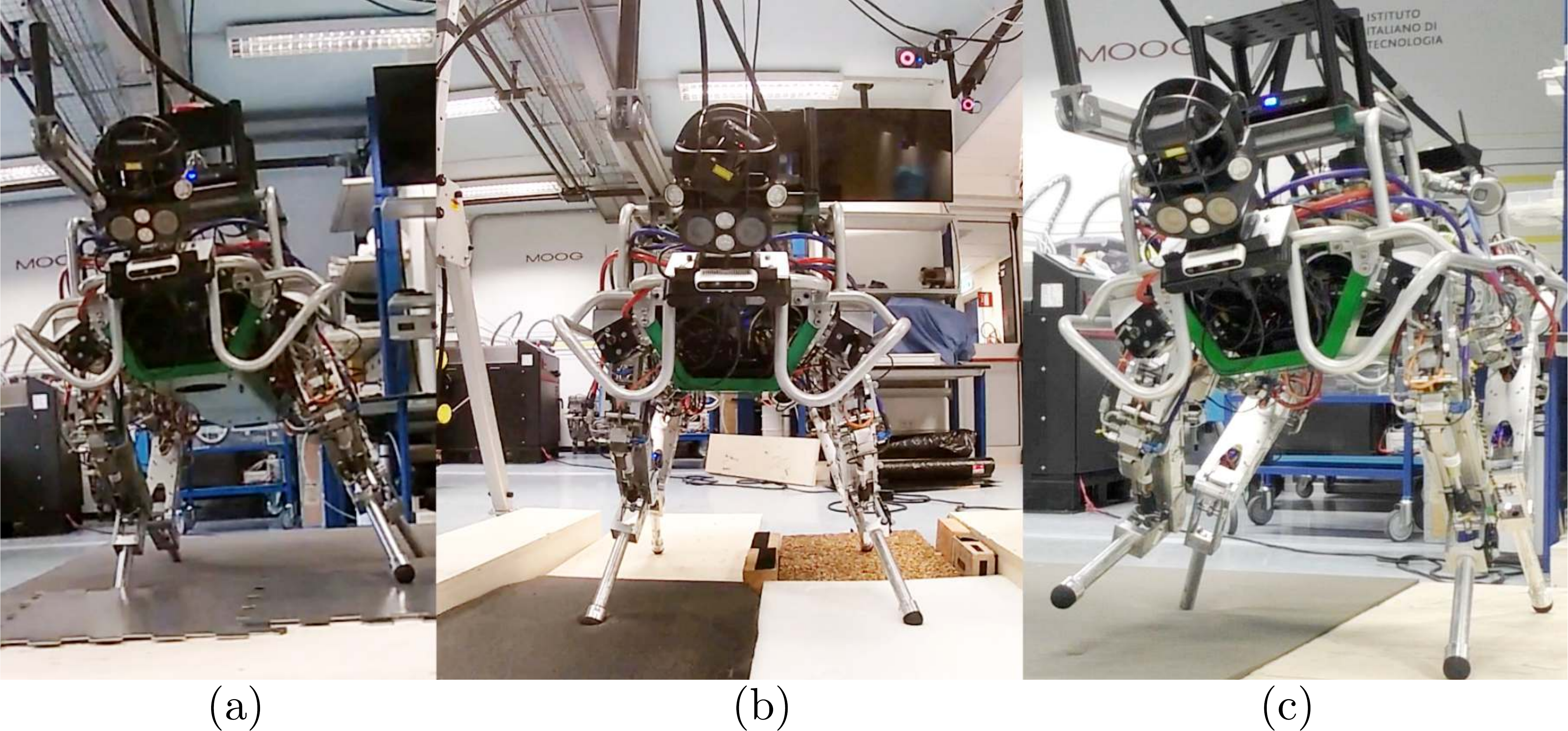}
	\caption{\footnotesize HyQ traversing multiple terrains of different compliances.}
	\label{fig:photos}
\end{figure}

To date, most of the work done on \gls{wbc} assumes that the ground is rigid 
(\ie rigid contact consistent).
However, if the robot traverses soft terrain (as shown in \fref{fig:photos}), 
the mismatch between the rigid assumption and the soft contact interaction can significantly affect the robot's 
performance and locomotion stability. 
This mismatch is due to the unmodeled contact dynamics between the robot and the terrain. 
In fact, under the rigid ground assumption, the controller can generate instantaneous changes to the 
\gls{grfs}. 
This is equivalent to thinking that  the terrain will respond with  an infinite bandwidth.

In order to robustly traverse a wide variety of terrains of different compliances, 
the \gls{wbc} must become \textit{\gls{c3}}. 
Namely, the \gls{wbc} should be terrain-aware. 
That said, a more general \gls{wbc} approach should be developed
that can adapt \textit{online} to the changes in the terrain compliance.

\subsection{\rev{Related Work - Soft Terrain Adaptation for Legged Robots}}
Locomotion over soft terrain can be tackled either from a control  or a planning  perspective.
In the context of locomotion control, 
Henze~\etaL~\cite{Henze2016} presented the first experimental 
attempt using a \gls{wbc} over soft terrain. 
Their~\gls{wbc} is based on the rigid ground 
assumption, but it allows for constraint relaxation.
This allowed the humanoid  robot TORO to adapt to a compliant surface. 
Their approach was further extended in~\cite{Henze2018} by dropping the rigid contact assumption and using an 
energy-tank approach. 
Despite balancing on compliant terrain, 
both approaches were only tested for one type of soft terrain when the robot was standing still. 

\rev{Similarly,} 
other works explicitly adapt to soft terrain by incorporating terrain knowledge (i.e., contact model) into 
their  balancing controllers. 
For example, Azad~\textit{et~al.}~\cite{Azad2015}  proposed a momentum based controller for balancing on soft terrain by 
relying on a nonlinear soft contact model. 
Vasilopoulos~\etaL~\cite{Vasilopoulos2018} proposed a similar hopping controller that models the 
terrain using a viscoplastic contact model.
However, these approaches were only tested in simulation and for monopods.

In  the context of locomotion planning, 
Grandia~\etaL~\cite{Grandia2019} indirectly adapted to soft terrain by
shaping the frequency of the cost function of their \gls{mpc} formulation.  
By penalizing high frequencies, they generated optimal motion plans that 
respect the bandwidth  limitations due to soft terrain. 
This approach was tested over three types of terrain compliances. 
However, it was not tested during transitions from one terrain to another. 
This approach showed
an improvement in the performance of the quadruped robot in simulation and experiment. 
However, the authors did not offer
the possibility 
to change their
tuning parameters online. 
Thus, they were not able to adapt the locomotion strategy  based on the compliance of the terrain. 

In contrast to the aforementioned work, other approaches relax the rigid ground assumption (hard contact constraint)
but not for soft terrain adaptation purposes. 
For instance, Kim~\etaL~\cite{Kim2019} 
implemented an approach to handle sudden changes in the rigid contact interaction. 
This approach relaxed the 
hard contact assumption in their \gls{wbc} formulation
by
penalizing the contact interaction in the cost function 
rather than incorporating it as a hard constraint. 
For  computational purposes, Neunert~\etaL~\cite{Neunert2018} and Doshi~\etaL~\cite{Doshi2019} 
proposed relaxing the rigid ground assumption.
Neunert~\etaL~used a soft contact model in their nonlinear \gls{mpc} 
formulation
to provide smooth gradients of the contact dynamics
to be more efficiently solved by their gradient based solver. 
The soft contact model did not have a physical meaning and the contact parameters were empirically chosen. 
Doshi~\etaL~proposed a similar approach which incorporates a 
slack variable that expands the feasibility region of the hard constraint.

Despite the improvement in performance of the legged robots over soft terrain in the aforementioned works,
none of them offered the possibility to adapt to the terrain \textit{online}.
Most  of the aforementioned works 
lack a general approach that can
deal with multiple terrain compliances 
or with transitions between them. 
Perhaps, one noticeable work (to date) in online soft terrain adaptation was proposed by 
Chang~\etaL~\cite{Chang2017}.
In that work, an iterative soft terrain adaptation approach was proposed. 
The approach relies on a non-parametric contact model 
that is simultaneously updated alongside an optimization based hopping controller.
The approach was capable of iteratively learning the terrain interaction and 
supplying that knowledge to the optimal controller. 
However, because the learning module was exploiting Gaussian process regression, which is computationally expensive,
the approach did not reach real-time performance and was only tested in simulation, for one leg, under one 
experimental condition (one terrain).
\subsection{\rev{Related Work - Contact Compliance Estimation in Robotics}} 
For contact compliance estimation, we need to accurately model the contact dynamics 
and estimate the contact parameters online.
In contact modeling, Alves~\etaL~\cite{Alves2015} presented a detailed overview of the types of parametric soft contact models used in the literature. 
In compliance estimation, Schindeler~\etaL~\cite{Schindeler2018} used a two stage polynomial identification 
approach to estimate the parameters of the Hunt and Crossley's~\gls{hc} contact model online.  
Differently, Azad~\etaL~\cite{Azad2016} used
a \gls{ls}-based estimation algorithm
and compared multiple contact models (including the \gls{kv} and the \gls{hc} models).
Other approaches that are not based on soft contact models
use force observers \cite{Coutinho2014} or neural networks \cite{Coutinho2013}. 
These aforementioned approaches in compliance estimation
were designed for robotic manipulation tasks.   

To date, the only  work on compliance estimation in 
legged locomotion was the one by Bosworth~\etaL~\cite{Bosworth2016}. 
The authors presented two online (in-situ) approaches to estimate 
the ground properties (stiffness and friction).
The results were promising and the approaches were validated on a quadruped robot 
while hopping over rigid and soft terrain. 
However, the estimated stiffness showed a trend, but was not accurate; 
the lab measurements of the terrain stiffness did not match the in-situ ones. 
Although the estimation algorithms could be implemented online, 
the robot had to stop to perform the estimation. 

\begin{figure}[tb]
	\centering
	\includegraphics[width=0.45\textwidth]{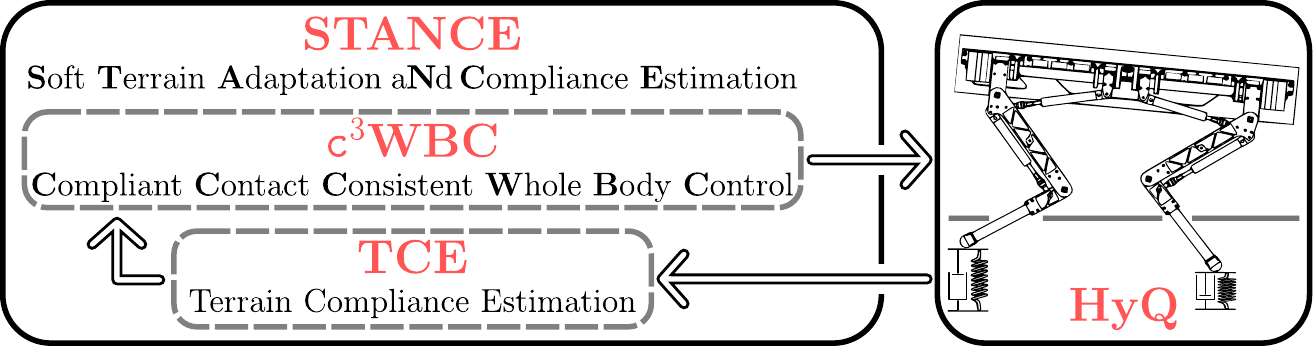}
	\caption{\footnotesize \rev{An overview of the} STANCE algorithm.}
	\label{fig:concept}
\end{figure}

\subsection{Proposed Approach and Contribution}
In this work, we propose an online soft terrain adaptation algorithm: \textbf{S}oft \textbf{T}errain \textbf{A}daptation a\textbf{N}d \textbf{C}ompliance 
\textbf{E}stimation~\gls{stance}. 
As shown in \fref{fig:concept}, \gls{stance} consists~of
\begin{itemize}[leftmargin=*]
\item A \gls{awbc} that is contact consistent to any type of terrain
\textit{given the terrain compliance}. 
This is done by extending the state-of-the-art \gls{wbc} in \cite{Fahmi2019}, hereafter denoted as the \gls{swbc}. 
In particular, \gls{awbc} incorporates a soft contact model into the \gls{wbc} formulation.
\item A \gls{ste} which is an online learning algorithm that 
provides the  \gls{awbc} with an estimate of the terrain compliance.
It is based on the same contact model that is incorporated in the \gls{awbc}.
\end{itemize}
The main contribution of \gls{stance} is that it can adapt to any type of terrain (stiff or soft)
online without pre-tuning. 
This is done by closing the loop of the \gls{awbc} with the \gls{ste}.
To our knowledge, this is the first implementation of such an approach in legged locomotion.

\gls{stance} is meant to overcome the limitations of the aforementioned approaches in soft terrain 
adaptation for legged robots. 
Compared to previous works on \gls{wbc} that tested 
their approach only during standing
\cite{Henze2016,Henze2018}, 
we test our \gls{stance} approach  during  locomotion. 
Compared to other approaches~\cite{Azad2015,Vasilopoulos2018}
that were tested on monopods in simulation, 
\gls{stance} is implemented and tested 
in experiment on \gls{hyq}.
Compared to previous work on soft terrain adaptation \cite{Grandia2019}, 
\gls{stance} can adapt to soft terrain \textit{online} and was tested on multiple 
terrains with different compliances and with transitions between them. 
Compared to \cite{Chang2017}, our \gls{ste} is computationally inexpensive, which allows \gls{stance} to
run real-time in experiments and simulations. 
Compared to the previous work done on compliance estimation, 
we implemented our \gls{ste} on a legged robot which is, to the  best of our knowledge, 
the first experimental validation of this approach. 
Differently from~\cite{Bosworth2016}, our \gls{ste} approach could be implemented
in parallel with any gait or task. 
We also achieved a more accurate estimation of the terrain
compliance compared to \cite{Bosworth2016}.

As additional contributions, 
we discussed the benefits (and the limitations) of exploiting the knowledge of the terrain in \gls{wbc}
based on the experience gained during extensive experimental trials. 
To our knowledge,  \gls{stance}  is the first work to present legged 
locomotion experiments crossing multiple terrains of different compliances.

\begin{figure*}[tb]
	\centering
	\includegraphics[width= \textwidth]{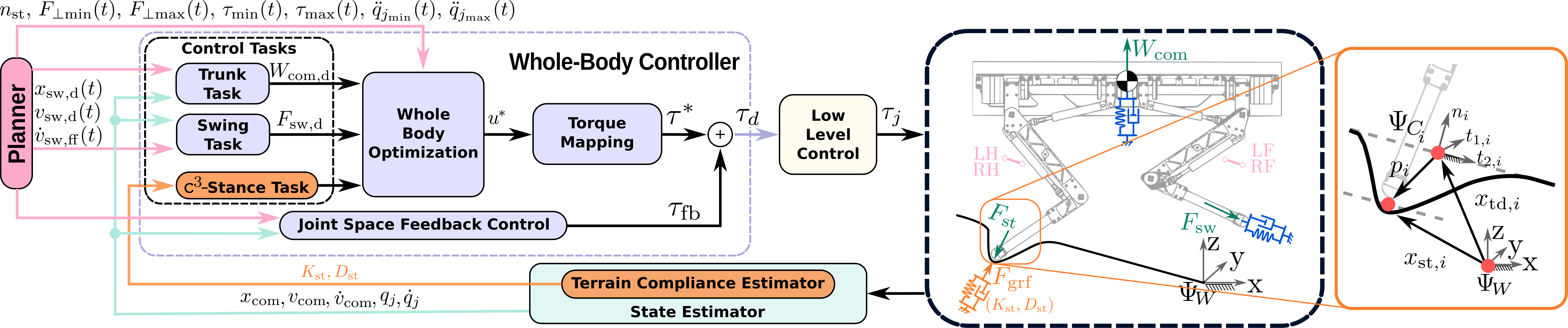}		
	\caption{\footnotesize Overview of the \gls{wbc} in	our locomotion framework.  
		The dashed black box presents the definition of the \gls{hyq}'s legs (\gls{lf}, \gls{rf}, \gls{lh} and \gls{rh}) 
		and the generated wrenches.
		The solid orange box presents the terminologies used for the soft contact model. 
		$\Psi_\mrm{W}$ 	is the world frame and 	$\Psi_{\mrm{C}_i}$ is the local contact frame for the leg $i$ \rev{fixed at the touch down position}.}
	\label{fig:blockDiagram}
\end{figure*}
\section{Robot model}
\label{secRobotModel}
Consider a legged robot with $n$ \gls{dofs} and $c$  feet.
The total dimension of the feet operational space $n_{\nmrm{f}}$ can be separated into stance  ($n_{\mrm{st}} = 
3 c_{\mrm{st}}$) and swing feet 
($n_{\mrm{sw}} = 3 c_{\mrm{sw}}$) where $c_{\mrm{st}}$ and $c_{\mrm{sw}}$ are the number of stance and 
swing legs respectively.
Assuming that all external forces are exerted on the stance feet,
the robot dynamics is written as

\begin{eqnarray}
\underbrace{ 
\mat{M_{\mrm{com}} & \vc{0}_{3\times 3} & \vc{0}_{3\times n} \\
	\vc{0}_{3\times 3} & \vc{M}_\theta & \vc{M}_{\theta j} \\
	\vc{0}_{n\times 3} & \vc{M}_{\theta j}^T & \vc{M}_j 
	}
}_{\vc{M}(\vc{q})}
\underbrace{
\mat{\ddot{x}_\mrm{com} \\ \dot{\omega}_{b}\\ \ddot{\vc{q}}_j}
}_{\ddot{\vc{q}}}
+ 
\underbrace{
\mat{ 
h_{\mrm{com}} \\ \vc{h}_\theta
	 \\ \vc{h}_j}
}_{h(q,\dot{q})} \nonumber \\ 
= \mat{ \vc{0}_{3\times 1} \\ \vc{0}_{3\times 1} \\ \vc{\tau_j}}
+ \underbrace{
\mat{\vc{J}_\mrm{st,com}^T \\ \vc{J}_{\mrm{st},\theta}^T \\ \vc{J}_{\mrm{st},j}^T}}
_{\vc{J}_\mrm{st}(\vc{q})^T}\vc{\grf}
\label{eq:full_dynamicsCOM}
\end{eqnarray}
\noindent
where $\vc{q}~\in~SE(3)~\times~\Rnum^n$ 
denotes the generalized robot states
consisting of 
the \gls{com} position $x_{\mrm{com}}$~$\in~\Rnum^3$,
the base orientation $R_b$~$\in~SO(3)$, and 
the joint positions $q_j$~$\in~\Rnum^n$.
The vector $\vc{\dot{q}}~=~[\dot{x}_\mrm{com}^T~\omega_b^T~\vc{\dot{q}}_j^T]^T~\in~\Rnum^{6+n}$
denotes the generalized velocities consisting of  
the velocity of the \gls{com} $\dot{x}_\mrm{com}$~$\in~\Rnum^3$, 
the angular velocity of the base $\omega_b$~$\in~\Rnum^3$, 
and the joint velocities $\dot{q}_j$~$\in~\Rnum^n$. 
The vector  $\ddot{\vc{q}}~=~[\ddot{x}_{\mrm{com}}^T~\dot{\omega}_{b}^T~\ddot{q}_j^T]^T~\in~\Rnum^{6+n}$ 
denotes the corresponding generalized accelerations.
All Cartesian vectors are expressed in the world frame $\Psi_W$ unless mentioned otherwise. 
%
%
$\vc{M}~\in~\Rnum^{(6+n) \times (6+n)}$ is the  inertia matrix.
$\vc{h}\in \Rnum^{6+n}$ is the force vector that accounts for Coriolis,
centrifugal, and gravitational forces.
$\vc{\tau_j}\in\Rnum^n$ are the actuated joint torques,
$\vc{\grf} \in\Rnum^{n_\mrm{st}}$ is the vector of \gls{grfs} (contact forces).
The Jacobian matrix $J$ $\in\Rnum^{n_\mrm{f} \times(6+n)}$
is separated into swing Jacobian $J_{\mrm{sw}}$~$\in\Rnum^{n_{\mathrm{sw}}~\times~(6+n)}$ and stance 
Jacobian  $J_{\mrm{st}}~\in~\Rnum^{n_{\mathrm{st}}~\times~(6+n)}$ which can be further expanded into 
$J_{\mathrm{st,com}}$  $\in\Rnum^{n_{\mathrm{st}} \times 3}$,
$J_{\mathrm{st,}\theta}$~$\in\Rnum^{n_{\mathrm{st}} \times 3}$,
and $J_{\mathrm{st},j}$  $\in\Rnum^{n_\mathrm{st} \times n}$.
The feet velocities $v~\in\Rnum^{n_\mrm{f}}$ are separated into stance $v_{\mrm{st}}~\in~\Rnum^{n_{\mrm{st}}}$ and swing $v_{\mrm{sw}}~\in~\Rnum^{n_{\mrm{sw}}}$ feet velocities. \rev{Similarly, the feet accelerations $\dot{v}~\in\Rnum^{n_\mrm{f}}$ are separated into stance $\dot{v}_{\mrm{st}}~\in~\Rnum^{n_{\mrm{st}}}$ and swing $\dot{v}_{\mrm{sw}}~\in~\Rnum^{n_{\mrm{sw}}}$ feet accelerations.}
The feet forces $F~=~[F_{\mrm{st}}^T \ 
F_{\mrm{sw}}^T]^T~\in~\Rnum^{n_\mrm{f}}$ are also separated into stance  $F_{\mrm{st}}~\in~\Rnum^{n_\mrm{st}}$ and swing 
$F_{\mrm{sw}}~\in\Rnum^{n_\mrm{sw}}$ feet forces.
We split the robot dynamics  \eref{eq:full_dynamicsCOM} into an unactuated floating base part (the first 6 rows) and an actuated part (the remaining $n$ rows) as
\begin{subequations}
	\begin{eqnarray}
	\vc{M}_u(\vc{q}) \ddot{\vc{q}} + h_u (q,\dot{q})  &=& \vc{J}_{\mrm{st},u}(\vc{q})^T\vc{\grf} \label{eq_unactuated}\\
	\vc{M}_a(\vc{q}) \ddot{\vc{q}} + h_j (q,\dot{q})  &=&  \vc{\tau_j} + \vc{J}_{\mrm{st},j}(\vc{q})^T\vc{\grf}  \label{eq_actuated}
	\end{eqnarray}
\end{subequations}
where $M_u$~$\in~\Rnum^{6\times6+n}$ and  $M_a$~$\in~\Rnum^{n\times6+n}$
are sub matrices of $M$, 
$h_u=[h_{\mrm{com}}^T~h_\theta^T ]^T$~$\in~\Rnum^{6}$ and
$h_j$~$\in~\Rnum^{n}$ are sub vectors of $h$, and
$J_{\mrm{st},u}=[J_{\mrm{st,com}}^T~J_{\mrm{st},\theta}^T]^T$.
Finally, we define the gravito-inertial wrench as
$W_\mrm{com}~=~\vc{M}_u(\vc{q}) \ddot{\vc{q}} + h_u (q,\dot{q})$ $\in \Rnum^6$.

\section{Standard Whole-Body Controller (sWBC)} 
\label{sec:wbc}
This section summarizes the \gls{swbc} as detailed in \cite{Fahmi2019}.
Besides the \gls{wbc}, our locomotion framework includes a locomotion planner, 
state estimator and a low-level
torque controller as shown in \fref{fig:blockDiagram}.
Given high-level user inputs, the planner generates the desired trajectories for the \gls{com}, trunk orientation and 
swing legs, and provides them to the \gls{wbc}. 
The state estimation provides the \gls{wbc} with the estimated states of the robot. 

The objective of the \gls{swbc} is to 
ensure the execution of the trajectories 
provided by the planner 
while keeping the robot balanced and reasoning about the robot's dynamics, 
actuation limits and the contact 
constraints \cite{Fahmi2019}. 
\rev{We denote the execution of the trajectories provided by the planner as \textit{control tasks}.
	These control tasks alongside the aforementioned constraints define the \gls{wbc} problem.}
The control problem is casted as a \gls{wbopt} problem via a \gls{qp} which solves for the optimal generalized 
accelerations 
and contact forces at each iteration of the  control loop~\cite{Herzog2016}.
The optimal solution of the \gls{wbc} is then mapped into joint torques 
that are sent to the low-level torque controller.

\subsection{Control Tasks}
\label{sec:controlTasks}
We categorize the \gls{swbc} \textit{control tasks} into: 
1) a \textit{trunk task} that tracks the desired trajectories 
of the \gls{com} position and trunk orientation, 
and 2) a \textit{swing task} that tracks the swing feet trajectories \cite{Fahmi2019}. 
Similar to a PD+ controller \cite{Ortega2013}, both tasks are achieved by a Cartesian-based impedance controller with a 
feed-forward term.
\rev{The feedforward terms are added in order to improve the tracking performance 
	of the tasks when following the trajectories from the planner \cite{Henze2016,Focchi2017}. }
The tracking of the trunk task  is obtained by the desired wrench at the \gls{com}  $W_{\mrm{com,d}} \in 
\Rnum^{6}$. 
This is generated by a Cartesian impedance at the \gls{com}, a gravity compensation 
term, and a feed-forward term.
Similarly, the tracking of the swing task can be obtained by the 
virtual force $F_{\mrm{sw,d}} \in \Rnum^{n_{\mrm{sw}}}$.
This is generated by a Cartesian impedance at the swing foot  
and a feed-forward term. 
As in~\cite{Fahmi2019}, we can also write the swing task at the acceleration level \rev{by defining the desired swing feet velocities $\dot{\vc{v}}_{\mrm{sw,d}}$ $\in \Rnum^{n_{\mrm{sw}}}$ as}
\begin{eqnarray}
\dot{\vc{v}}_{\mrm{sw,d}} = \dot{\vc{v}}_{\mrm{sw,ff}} +
\mrm{K}_{\mrm{sw}} \Delta x_{\mrm{sw}} +
\mrm{D}_{\mrm{sw}} \Delta v_{\mrm{sw}} 
\label{eq:swingTaskAcc}
\end{eqnarray}
where $\mrm{K}_{\mrm{sw}}, \mrm{D}_{\mrm{sw}} \in\Rnum^{n_\mrm{sw}\times n_\mrm{sw}}$
are positive definite PD gains,
$\Delta x_{\mrm{sw}} = x_{\mrm{sw,d}} - x_{\mrm{sw}} $ $\in \Rnum^{n_{\mrm{sw}}}$ and $\Delta v_{\mrm{sw}} = v_{\mrm{sw,d}} - 
v_{\mrm{sw}}$ $\in \Rnum^{n_{\mrm{sw}}}$ are tracking 
errors of the swing foot position and velocity, respectively,
and $\dot{\vc{v}}_{\mrm{sw,ff}}$ is a feed-forward term.

\subsection{Whole-Body Optimization} 
\label{sec:wbopt}
\rev{To accomplish the \gls{swbc} objective (the control tasks in  \sref{sec:controlTasks} and constraints), 
we formulate the \gls{wbopt} problem presented in \defref{eqn_wbopt} and detailed in \cite{Fahmi2019}.}
\subsubsection{\rev{Decision Variables}}
As shown in \defref{eqn_wbopt}, we choose the generalized 
accelerations~$\ddot{\vc{q}}$ and the contact forces~$\grf$ as the decision variables~$\vc{u}~=~[\ddot{\vc{q}}^T~\vc{\grf}^T]^T~\in~\Rnum^{6+n+n_{\mrm{st}}}$. 
Later in this subsection, we will augment the vector of decision 
variables with a slack term  $\eta \in \Rnum^{n_{\mrm{sw}}}$.

\subsubsection{\rev{Cost}}
\rev{The cost function  \eref{wbopt_cost_function} consists of two terms. 
The first term ensures the tracking of the trunk task by minimizing the two-norm of the tracking error between the actual $W_{\mrm{com}}$ 
and desired  $W_{\mrm{com,d}}$ \gls{com} wrenches.
The second term in  \eref{wbopt_cost_function}
regularizes the solution and penalizes the slack variable.}

\subsubsection{\rev{Physical Consistency}}
The equality constraint \eref{wbopt_physical_consistency} enforces the physical consistency between $\grf$ and 
$\ddot{\vc{q}}$ by \rev{ensuring} that the contact wrenches due to $\grf$ will sum up to $W_\mrm{com}$. 
\rev{This is done by imposing \frev{the unactuated dynamics \eref{eq_unactuated}} as an equality
constraint.}

\subsubsection{\rev{Stance Task}}
To remain  \textit{contact consistent}, we incorporate the \textit{stance task}
that enforces the stance legs to remain in contact with the terrain. 
Since the \gls{swbc} is assuming a rigid terrain, 
the stance feet are forced to remain stationary
in the world frame,  i.e.,~$v_{\mrm{st}} = \dot{v}_{\mrm{st}} = 0$
(see~\cite{Fahmi2019}). 
As a result, we incorporate the rigid contact model
in the \gls{swbc} formulation as an equality constraint at the acceleration level \eref{wbopt_rigid} in
order to have a direct dependency on the decision variables. 
\rev{In detail,  since $v_{\mrm{st}} = J_{\mrm{st}} \dot{q} = 0$, differentiating once with respect to time yields
	$\dot{v}_{\mrm{st}} = \vc{J}_{\mrm{st}}\ddot{\vc{q}}+\dot{\vc{J}}_{\mrm{st}}\dot{\vc{q}}=0$. }

\subsubsection{\rev{Friction and Normal Contact Force}}
The inequality constraint \eref{wbopt_friction} enforces the friction constraints \rev{by ensuring that the contact forces lie inside the  friction cones}.
\rev{This is done by limiting the tangential component of the  \gls{grfs}~$\grfp{\parallel}$.}
The inequality constraint \eref{wbopt_contact_force} enforces  constraints 
on the normal component of the \gls{grfs}~$\grfp{\perp}$. 
This includes the unilaterality constraints which encodes that the legs can 
only push on the ground  by setting an  ``almost-zero'' lower bound $F_{\mrm{\min}}$ to 
$\grfp{\perp}$.
They also allow a smooth loading/unloading of the legs, and set
a varying upper bound $F_{\mrm{\max}}$ to   $ \grfp{\perp}$. 
\rev{For the detailed implementation of the inequality constraints~\eref{wbopt_friction}~and~\eref{wbopt_contact_force}, refer to \cite{Focchi2017}.}

\subsubsection{\rev{Swing Task}}
We implement the tracking of the swing task  (in \sref{sec:controlTasks})
at the acceleration level  \eref{eq:swingTaskAcc} rather than the force level
since we can express \rev{the swing feet velocities} $\dot{v}_{\mrm{sw}}$ as a function of  $\ddot{q}$ which is a decision variable, 
\ie~$\dot{\vc{v}}_{\mrm{sw}}(\vc{q})= \vc{J}_{\mrm{sw}}\ddot{\vc{q}} + 
\dot{\vc{J}}_{\mrm{sw}}\dot{\vc{q}}$. 
This task could be encoded as an equality constraint
$\dot{v}_{\mrm{sw}} =\dot{v}_{\mrm{sw,d}}$. 
Yet, it is important to relax this hard constraint 
when the joint kinematic limits are reached \rev{(see~\cite{Fahmi2019})}. 
Hence, the swing task is encoded in \eref{wbopt_swing_task} 
by an inequality constraint that bounds the solution around the original hard constraint
and a slack term  $\eta$ that is penalized for its non-zero values in the cost function \eref{wbopt_cost_function}
and is constrained to remain non-negative in \eref{wbopt_swing_task}.

\subsubsection{\rev{Torque and Joint Limits}}
The torque and joint limits  are enforced in the 
inequality constraints \eref{wbopt_torque_limits}  and  \eref{wbopt_joint_limits}, respectively.

\subsubsection{\rev{Torque Mapping}}
The \gls{wbopt} \eref{wbopt_cost_function}-\eref{wbopt_rigid},~\eref{wbopt_friction}-\eref{wbopt_joint_limits} generates
optimal joint accelerations $\ddot{\vc{q}}_j^*$  and contact forces 
$\vc{\grf}^*$, that are mapped into \rev{optimal} joint torques $\tau^*$ 
and sent to the low-level controller
using \frev{the actuated dynamics~\eref{eq_actuated}~as}
\frev{\begin{equation}
\tau^{*} = M_a\ddot{q}^* + h_j - J_{\mrm{st},j}^T \grf^*
\label{eq:torques}
\end{equation}}

\begin{definition}
	\centering
	\caption{Whole-Body Optimization: \gls{swbc} Vs. \gls{awbc}}\label{eqn_wbopt}
	\vspace{-0.9cm}
	\[
	\begin{minipage}[t]{0.15 \linewidth}
	\begin{align*}
	~&\text{\footnotesize (Trunk Task)}\\[4pt]
	~&\text{\footnotesize (Decision Variables)}\\
	~&~\\
	~&\text{\footnotesize(Physical Consistency)}\\[4pt]
	~&\text{\footnotesize\cancel{(Stance Task)}}\\
	~&\text{\footnotesize\textcolor{RoyalBlue}{\textbf{(\texttt{c}$^3$-Stance Task)}}}\\ ~&\\ ~&\\[2pt] 
	~&\text{\footnotesize(Friction)}\\[1pt]
	~&\text{\footnotesize(Normal Contact Force)}\\[1pt]
	~&\text{\footnotesize(Swing Task)} \\[0pt]
	~&\text{\footnotesize(Torque Limits)} \\
	~&\text{\footnotesize(Joint Limits)} 
	\end{align*}		
	\end{minipage}
	\begin{minipage}[t]{0.65\linewidth}
	\begin{gather} 
	\underset{\vc{u}} {\text{min~}}  
	\rev{\Vert W_{\mrm{com}}-W_{\mrm{com,d}} \Vert^2_Q + \bmcolor{\Vert\vc{u}\Vert^2_{R}}}
	\label{wbopt_cost_function} \\
	\vc{u}=[\ddot{\vc{q}}^T \ \vc{\grf}^T  \ \vc{\eta}^T \ \bmcolor{\vc{\epsilon}^T }]^T \nonumber 
	\label{wbopt_dec_variables}\\
	\text{s.t.:}   \hspace{135pt}  \nonumber  \\
	\frev{M_u \ddot{\vc{q}} + h_u = J^T_{\mrm{st},u}  \grf  }
	\label{wbopt_physical_consistency}\\
	\cancel{\dot{v}_{\mrm{st}}=\vc{J}_{\mrm{st}}\ddot{\vc{q}}+\dot{\vc{J}}_{\mrm{st}}\dot{\vc{q}} = 
		0} \label{wbopt_rigid}\\
	\bmcolor{\grf = K_{\mrm{st}} \epsilon   + D_\mrm{st} \dot{\epsilon}  } 
	\label{wbopt_fgrf_epsilon}\\
	\bmcolor{\dot{v}_{\mrm{st}} =  
		\vc{J}_{\mrm{st}}\ddot{\vc{q}}+\dot{\vc{J}}_{\mrm{st}}\dot{\vc{q}} = 
		- \ddot{\epsilon} } 
	\label{wbopt_ddot_epsilon}\\
	\bmcolor{ \epsilon \geq 0  } \label{wbopt_penetration_slack} \\
	\vert \grfp{\parallel} \vert \leq \mu |\grfp{\perp} | \label{wbopt_friction} \\
	F_{\mrm{\min}} \leq \grfp{\perp} \leq F_{\mrm{\max}} \label{wbopt_contact_force}\\
	-\eta\leq \dot{v}_{\mrm{sw}} - \dot{\vc{v}}_{{\mrm{sw,d}}}  \leq\eta ,~ \eta  \geq 0  
	\label{wbopt_swing_task}\\
	\vc{\tau}_{\min} \leq \vc{\tau}_j \leq \vc{\tau}_{\max}  \label{wbopt_torque_limits} \\
	\ddot{\vc{q}}_{j_{\min}} \leq \ddot{\vc{q}}_j \leq \ddot{\vc{q}}_{j_{\max}}  
	\label{wbopt_joint_limits}
	\end{gather}
	\end{minipage}
	\nonumber
	\]
	\vspace{-0.2cm}
	\setcounter{algorithm}{0}
\end{definition}

\subsection{\rev{Feedback Control}}
\rev{The computation of the optimal torques  $\tau^*$ relies on the inverse dynamics 
in \eref{eq:torques} which might be prone to model inaccuracies \cite{Righetti2013}. 
In order to tackle this issue, the desired torques $\tau_d$ sent to the lower level 
control could combine the optimal torques $\tau^*$ in  \eref{eq:torques}
with a feedback controller $\tau_{\mrm{fb}}$ as shown in \fref{fig:blockDiagram}. 
The feedback controller improves the tracking performance if the dynamic model of the robot becomes less accurate \cite{Righetti2013}. 
The feedback controller is a proportional-derivative (PD) joint space impedance controller \cite{Focchi2017}. 
}

\begin{assumpB}
\rev{Throughout this work and similar to \cite{Fahmi2019}~and~\cite{Herzog2016}, we found it sufficient to use only the inverse-dynamics term  
(the optimal torques $\tau^*$) and not the joint feedback part. 
This is due to the fact that we can identify the parameters 
of our dynamic model with sufficient accuracy as detailed in~\cite{Tournois2017}. 
That said, we carried out the simulation and experiment without any need of the feedback loop. }
\end{assumpB}

\section{C$^3$ Whole Body Controller} \label{wbc:soft}
Over soft terrain, the feet positions are non-stationary  and are allowed to deform the terrain. 
Thus, the rigid contact assumption of the stance task \eref{wbopt_rigid}
in the \gls{swbc} does not hold anymore and should be dropped.
To be \gls{c3}, the interaction between the stance feet 
and the soft terrain must be governed not just by  the robot dynamics but also by 
the soft contact dynamics. 
That said, the \gls{awbc} extends the \gls{swbc} by: 1)~modeling the soft contact dynamics  
and incorporating it as a \textit{stance task} similar to the control tasks in \sref{sec:controlTasks},
and 2)~encoding the stance task in the  \gls{wbopt} as a function of the decision
variables. 
The differences between the \gls{swbc} and the \gls{awbc} are highlighted in 
\textbf{\textcolor{RoyalBlue}{boldface}} in 
\defref{eqn_wbopt}.
\begin{assumpB}
\rev{The term \textit{contact consistent} is a well-established term in the literature that 
was initially introduced in \cite{JaeheungPark2006}. 
It implies formulating the control structure to account for the contact with the environment.
The term \gls{c3} is an extension of the term contact consistent. 
Hence, \gls{c3} implies formulating the control structure to account for the \textit{compliant} contact with the 
environment.}
\end{assumpB}

\subsection{\gls{c3}-Stance Task}
\label{sec:contactmodelling}
We model the soft contact dynamics with a simple explicit model (the \gls{kv} model). 
This consists of 3D linear springs and dampers 
normal and tangential to the contact point \cite{Neunert2018}. 
The normal direction of this impedance 
emulates the normal terrain deformation while 
the tangential ones emulate the shear deformation.  
Although several models that accurately emulate contact dynamics are available
\cite{Azad2010,Ding2013,Alves2015}, we implemented the \gls{kv} model for several reasons. 
First, since the model is linear in the parameters, it fits our \gls{qp} formulation. 
Second, estimating the parameters of the \gls{kv} model is computationally efficient. 
As a result, using this model,  we can run a learning algorithm online which would be challenging
if a model similar to \cite{Chang2017} is used. 
For a legged robot with point-like feet,  
for each stance leg $i$, we formulate the contact model in the world frame as
\begin{eqnarray}
\grfp{i} =    k_{\mrm{st},i}  p_i  + d_{\mrm{st},i} \dot{p}_i
\end{eqnarray}
\rev{where $k_{\mrm{st},i} \in \Rnum^{3\times3}$,
$d_{\mrm{st},i} \in \Rnum^{3\times 3}$,
$\grfp{i} \in \Rnum^3$, $ p_i \in \Rnum^3$, and $ \dot{p}_i  \in \Rnum^3$ 
are the terrain stiffness, the terrain damping,
the \gls{grfs}, the penetration 
and the penetration rate of the $i$-th stance leg, all expressed in the world frame, respectively (see \fref{fig:blockDiagram}).}
\rev{We define} $ p_i$ and $ \dot{p}_i$ as
\begin{eqnarray}
p_i =  x_{\mrm{td},i} - x_{\mrm{st},i} \text{,} \hspace{5em } \dot{p}_i = 0 - v_{\mrm{st},i}
\label{eq:penetrationDef}
\end{eqnarray}
where  $x_{\mrm{td},i} \in \Rnum^3$ denotes the position of 
the contact point of foot $i$ at the touchdown \rev{in the world frame}. 
By appending all of the stance feet, the contact model can be re-written in a compact form as
\begin{equation}
\grf = K_{\mrm{st}} p + D_{\mrm{st}} \dot{p} =  K_{\mrm{st}} (x_{\mrm{td}} - x_{\mrm{st}}) -  D_{\mrm{st}} v_{st}
\label{contactmodel1}
\end{equation}
where $K_{\mrm{st}}$ $\in R^{n_{\mrm{st}} \times n_{\mrm{st}}}$ and $D_{\mrm{st}}$ $\in R^{n_{\mrm{st}} \times 
n_{\mrm{st}}}$ are the block-diagonal  stiffness and 
damping matrices of the terrain of all the stance feet, respectively,
and $x_{\mrm{td}}$ $\in \Rnum^{n_{st}}$ are the touchdown positions of all the stance feet.

Similar to \sref{sec:controlTasks}, we deal with the contact model~\eref{contactmodel1} as 
another \gls{wbc} task (alongside the trunk and swing \eref{eq:swingTaskAcc} tasks). 
We can think of  \eref{contactmodel1} 
as a desired stance task that keeps the \gls{wbc} \gls{c3}. 
This stance task  is achieved by a Cartesian impedance at 
the stance foot which is represented by the 
impedance of the terrain ($K_{\mrm{st}}$  and $D_{\mrm{st}}$). 
This similarity makes us encode the contact model in the \gls{wbopt} as 
a stance constraint similar to what we 
did for the swing task in \sref{sec:wbopt}.
Hereafter, we refer to this stance task as the~\gls{c3}-stance task (see \fref{fig:blockDiagram}).

\subsection{Whole-Body Optimization Revisited}
The \gls{c3}-stance task is included in the \gls{wbopt}  by writing the soft contact model 
\eref{contactmodel1} as a function of the decision variables. 
Ideally, we can directly reformulate  \eref{contactmodel1} 
as a function of $\grf$ and $\dot{v}_{\mrm{st}}$.
Indeed, $\dot{v}_{\mrm{st}}$ can be expressed as a function of the 
joint accelerations $\ddot{q}$ which is a decision variable (as explained in~\cite{Fahmi2019}).  
By numerically integrating $\dot{v}_{\mrm{st}}$ 
(once to obtain $v_{\mrm{st}}$ and twice to obtain $x_{\mrm{st}}$),
we can associate $\grf$ with $\dot{v}_{\mrm{st}}$.
This approach requires the knowledge of  $x_{\mrm{td}}$  to compute $p$ 
which might be prone to estimation errors and it requires  a reset  of the integrator at every touchdown.

We choose a more convenient approach which  
is to add the \textit{desired} foot penetration $\epsilon$
as an extra decision variable  in the \gls{wbopt} formulation. 
The difference between  $p$ and $\epsilon$ is that~$p$ is
the \textit{actual} 	penetration due to the interaction with the soft contact
while $\epsilon$ is the \textit{desired} penetration \rev{in the world frame} generated from the optimization problem. 
Both variables imply the same physical phenomenon (the soft contact deformation). 
That said, 
we can rewrite $\grf$ in \eref{contactmodel1} 
as a function of $\epsilon$ and $\dot{\epsilon}$ (by numerically differentiating $\epsilon$)
without the previous knowledge of $x_{\mrm{td}}$, which is advantageous. 

To do so, 
$\epsilon$ is appended to the vector of decision variables~$u$ and regularized in \eref{wbopt_cost_function}.
Then, we incorporate \eref{contactmodel1} directly as a function of  $\grf$ and $\epsilon$ 
as in the equality constraint~\eref{wbopt_fgrf_epsilon}.
We numerically differentiate $\epsilon$ to obtain 
$\dot{\epsilon}_k = \frac{\epsilon_k - \epsilon_{k-1}}{\Delta t}$. 
To maintain physical consistency, we need to enforce an additional
constraint 
between the desired penetration $\epsilon$ and
the contact acceleration
($\ddot{\epsilon} = -\dot{v}_{\mrm{st}}$). 
This is encoded as an equality constraint as shown in \eref{wbopt_ddot_epsilon}. 
To do so, we numerically differentiate 
$\epsilon$ twice to obtain $\ddot{\epsilon}_k = \frac{\epsilon_k - 2\epsilon_{k-1} + \epsilon_{k-2}}{\Delta t^2}$. 
We also ensure the consistency of the physical contact model throughout the optimization
problem by ensuring that the penetration is always positive in 
\eref{wbopt_penetration_slack}. 

We also consider
the loading and unloading phase, explained in \cite{Fahmi2019}
 and \cite{Focchi2017}, to be terrain-aware. 
We tune the loading and unloading phase period  $T_{l/u}$ for each leg
to follow the settling time of a second order system response
that is a function of the terrain compliance
and the robot's mass \cite{Franklin2014}. 
Hence, $T_{l/u}$ is
\begin{equation}
	T_{l/u} = 4.6 / \sqrt{\frac{k_{\mrm{st,i}}}{m_e}}
\end{equation}
where $m_e$ is the equivalent mass felt at the robot's 
feet (\ie the weight of the robot $m_{R}$ spread 
across its stance feet~$m_e~=~m_{R} / n_{\mrm{st}}$)
and the constant in the numerator represents a $1\%$ steady-state error.

Finally,
the \gls{wbopt} 
\eref{wbopt_cost_function}, \eref{wbopt_physical_consistency}, \eref{wbopt_fgrf_epsilon}-\eref{wbopt_joint_limits} 
generates
optimal joint accelerations $\ddot{\vc{q}}_j^*$  and contact forces 
$\vc{\grf}^*$, that are  mapped into \rev{optimal} joint torques $\tau^*$ 
and sent to the low-level controller
using the actuated part of the 
robot's dynamics as shown in~\eref{eq:torques}. 
\rev{Note that similar to the \gls{swbc}, we  found it sufficient to use only the inverse-dynamics term  
	(the optimal torques $\tau^*$) and not the joint feedback part. }

As explained above, adding $\epsilon$ as a decision variable
involved adding  two constraints in the optimization
which increases the problem size and the computation time. 
Yet, we are still able to run the \gls{awbc} in real-time.  
The advantage of our approach is that the knowledge of the 
touchdown position  $x_{\mrm{td}}$ is not required.
We only need the previous two time instances of the penetration $\epsilon_{k-1}$
and $\epsilon_{k-2}$ that we already computed in the previous control loops.

\section{Terrain Compliance Estimation} 
\label{sec:softterrainestimation}
\begin{figure*}[!t]
	\centering
	\includegraphics[width=0.98\textwidth]{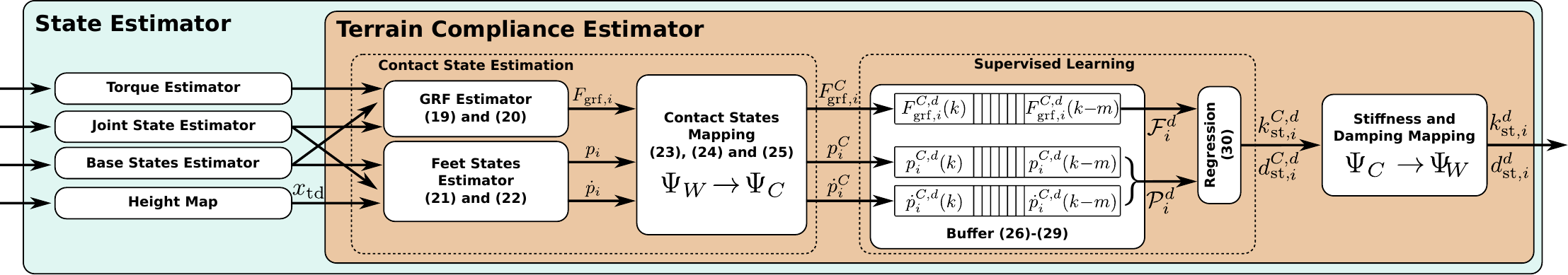}
	\caption{\footnotesize \rev{Overview of the \gls{ste}'s architecture inside the state estimator.}}
	\label{tce_arch}
\end{figure*}
The purpose of the \gls{ste} is to estimate \textit{online} the terrain parameters 
(namely $K_{\mrm{st}}$ and $D_{\mrm{st}}$) based on the states of the robot. 
\rev{It is a stand-alone algorithm that is decoupled from the \gls{awbc}.}
\rev{The \gls{ste} uses the contact model \eref{contactmodel1}.
	Based on that, the current measurement of the
	\textit{contact states} (contact status $\alpha$, \gls{grfs} $\grf$, the penetration $p$, and the penetration rate $\dot{p}$) of each leg $i$ at every time step are required.}
\rev{Given the contact states, we use supervised learning to learn the terrain parameters.}
\rev{As shown in \fref{tce_arch}, the \gls{ste} consists of two main modules: 
	contact state estimation (\sref{sec_contact_state})
	and supervised learning (\sref{sec_supervised_learning}). 
	The contact state estimation module estimates the contact states and provides it to the supervised learning
	module that collects these data and computes the estimates of the terrain parameters.}

\subsection{\rev{Contact State Estimation}}
\label{sec_contact_state}
\rev{The contact states are estimated solely from the current states of the robot
	by the state estimator.
	The \grfs are estimated from the torques and the joint states, and
	the penetration and its rate are estimated from the floating base (trunk) states and the joint states.} 
\subsubsection{GRFs Estimation}
\rev{To estimate the GRF, we use \frev{actuated part of the dynamics in \eref{eq_actuated}} as
\frev{
\begin{equation}
F_{\mrm{grf},i} = \alpha_i \vc{J}_{j,i}^{-T} (M_{a,i}\ddot{\vc{q}}_{i} + \vc{h}_{j,i} - \vc{\tau}_{j,i})
\label{grf_est}
\end{equation}
where} $F_{\mrm{grf},i}$, $\vc{J}_{j,i}$, \frev{$M_{a,i}$, $\ddot{q}_{i}$},  $\vc{h}_{j,i}$, and $\vc{\tau}_{j,i}$
correspond to 
$F_{\mrm{grf}}$, $\vc{J}_{\mrm{st},j}$, \frev{$M_a$, $\ddot{q}$}, $\vc{h}_j$, and $\vc{\tau}_j$ 
for the $i$-th leg, respectively. Additionally,  $\alpha_i$ is the contact status
variable that detects if there is a contact in the $i$-th leg or not. 
The contact is detected when the GRF exceed a certain threshold $F_{\min}$.
Hence, $\alpha_i$ computed along the normal direction of the $i$-th leg $n_i$ as:
\frev{\begin{equation}
\alpha_i =\begin{cases}
1, & \text{if $n_i^T (\vc{J}_{j,i}^{-T} (M_{a,i}\ddot{\vc{q}}_{i} + \vc{h}_{j,i} - \vc{\tau}_{j,i}) \geq F_{\min}$}\\
0, & \text{otherwise}
\end{cases}
\label{contact_status}
\end{equation}
}
}

\subsubsection{Penetration Estimation}
\rev{
	As shown in \eref{eq:penetrationDef}, we estimate the penetration and its rate
	using the stance feet positions $x_{\mrm{st},i}$ and velocities $v_{\mrm{st},i}$,
	and the touchdown position $x_{\mrm{td},i}$ all in the \textit{world} frame.
	To estimate the feet states in the world frame, we use the forward kinematics and the base state in the world frame.
	Thus, the penetration and its rate are written as}
\rev{\begin{eqnarray}
	p_i &=& x_{\mathrm{td},i} - x_{\mrm{st},i}
	=x_{\mathrm{td},i} - x_{b} -  R_B^W x_{\mrm{st},i}^B
	\label{p_est}\\
	\dot{p}_i &=& - v_{\mrm{st},i}
	= - v_b - R_B^W v_{\mrm{st},i}^B - (\omega_b \times R_B^W) x_{\mrm{st},i}^B
	\label{dotp_est}
	\end{eqnarray}
	where
	$x_{b}$ $\in \Rnum^3$ and
	$v_b$ $\in \Rnum^3$
	are the base position and velocity in the world frame, respectively. 
	The terms
	$x_{\mrm{st},i}^B$ and
	$v_{\mrm{st},i}^B$
	are the stance feet position and velocity of the $i$-th leg in the base frame, respectively. 
	The terms
	$R_B^W$ $\in SO(3)$ and 
	$\omega_b$ are the rotation matrix mapping vectors from the base frame to 
	the world frame and the base angular velocity, respectively.
	$
	$
	The  touch down positions are  obtained using a height map.}

\subsubsection{Contact States Mapping}
	\rev{Since, the \gls{kv} model consists of 3D linear springs and dampers,  
		normal and tangential to the contact point, this makes the stiffness and damping 
		matrices diagonal with respect to the contact frame. 
		However,  if expressed in the world frame, the 
		stiffness and damping matrices become dense.
		Thus, if we formulate the \gls{kv} model  in the contact frame rather than the world frame,
		we estimate less number of elements per matrix per leg: three elements instead of nine.
		Henceforth, the \gls{kv} model in the \gls{ste} should be formulated with respect to the contact frame
rather than the world frame to reduce the computational complexity.}
\rev{To do so, the \grfs \eref{grf_est}, the penetration \eref{p_est}  and its rate \eref{dotp_est} 
	of the $i$-th leg 
	are transformed from the world frame $\Psi_W$ to the contact frame $\Psi_{C_i}$ as (see \fref{fig:blockDiagram})
\begin{eqnarray}
F_{\mathrm{grf},i}^C &=& R_W^{C_i} \grfest{i} \label{grf_trans}\\
p_{i}^C &=&  R_W^{C_i} p_i \label{p_trans}\\
\dot{p}_{i}^C &=& R_W^{C_i} \dot{p}_i \label{pdot_trans}
\end{eqnarray}
where the superscript $\bullet^C$ refers to the contact frame and~$R_W^{C_i}$ is the rotation matrix mapping from the world $\Psi_W$ to the contact~$\Psi_{C_i}$ frames for the $i$-th leg. 
Note that the transformation~\eref{pdot_trans} 
is linear
since the contact frame is \textit{fixed} with respect to the world frame at the touch down position (\ie $\dot{R}_W^{C_i} = 0$).}

\subsection{\rev{Supervised Learning}}
\label{sec_supervised_learning}
\rev{Considering the contact model in the contact frame
and using the estimated contact states  \eref{grf_trans}-\eref{pdot_trans},}
we learn the terrain parameters online via supervised learning.
In particular, we use weighted linear least squared regression.
The algorithm is treated as a batch algorithm with $m$-examples such that, 
at every time instant $k$, we gather samples from the 
previous $m$ time instances and 
compute the terrain parameters
\cite{Stulp2015}.

\rev{For the $k$-th time instant,
	of the $i$-th leg in the $d\text{-th}$ direction ($d \in \{n_i,t_{1,i},t_{2,i}\}$, see \fref{fig:blockDiagram}),
	the terms $F_{\mathrm{grf},i}^{C,d}(k)$, $p_{i}^{C,d}(k)$, and $\dot{p}_{i}^{C,d}(k)$
	are estimated as shown in \sref{sec_contact_state} where~$\bullet_i^d(k)$ refers to the $k$-th time instance of the $i$-th leg in the $d$-th direction.}
\rev{That said, we construct the following objects (buffers) with size $m$
\begin{eqnarray}
\mathcal{F}_i^d &=& \mat{ F_{\mathrm{grf},i}^{C,d}(k) & \cdots &F_{\mathrm{grf},i}^{C,d}(k-m) }^T \label{tce1}\\
\mathrm{P}_i^d &=& \mat{p_{i}^{C,d}(k) & \cdots & p_{i}^{C,d}(k-m) }^T \label{tce2}\\
\dot{\mathrm{P}}_i^d &=& \mat{ \dot{p}_{i}^{C,d}(k) & \cdots & \dot{p}_{i}^{C,d}(k-m) }^T \label{tce3}\\
\mathcal{P}_i^d &=& \mat{\mathrm{P}_i^{d}&  \dot{\mathrm{P}}_i^{d}} \label{tce4}
\end{eqnarray}
where $\mathcal{F}_i^d \in \Rnum^{m}$
is the \gls{grfs} buffer and $\mathcal{P}_i^d \in \Rnum^{m \times 2}$
is the penetration, and penetration rate buffer.
Given $\mathcal{F}_i^d$ and  $\mathcal{P}_i^d$ as inputs and outputs of the learning algorithm respectively, 
we estimate the terrain impedance parameters as
$I_i^d   =\mat{k_{\mathrm{st},i}^{C,d} ~ d_{\mathrm{st},i}^{C,d}}^T \in \Rnum^{2}$
using the analytical solution
\begin{equation}
	I_i^{d}   =  (\mathcal{P}_i^{dT} W \mathcal{P}_i^d  )^{-1} \mathcal{P}_i^{dT} W  \mathcal{F}_i^d
	\label{tce5}
\end{equation}
where $k_{\mathrm{st},i}^{C,d}\in \Rnum$ and $d_{\mathrm{st},i}^{C,d} \in \Rnum$ are the terrain stiffness and damping parameters expressed in the contact frame.}
The matrix $W \in \Rnum^{m \times m}$ is a weighting matrix used to 
penalize the error on most recent sample compared to the less 
recent ones and thus, giving more importance to  the most recent samples. 

\rev{All of the legs in the learning algorithm are decoupled.} 
We found it advantageous to treat each leg separately because the robot can 
be  standing on a different terrain at each foot.

\subsection{\rev{Implementation Details}}
\rev{\assref{tce_alg} sketches the entire \gls{ste} process}.
\rev{To initialize the buffers, we acquire samples when the robot is at full stance
	and return the first estimate of $I_i^d$ once the buffers are full.
	After initialization, 
	we acquire samples and update the buffers only when the leg is at stance.} 

\rev{The buffers are continuously updated in a sliding window fashion. 
	When a leg finishes the swing phase and is at a new touch down,
	it continues to use the previous samples from the previous stance phase.  This is advantageous since it gives a smooth transition between terrains, but it adds a delay. }
\begin{assumpB}
	\rev{Since the \gls{awbc} formulation is based in the world frame, it is essential 
		to map the estimated stiffness and damping matrices
		back to the world frame before providing them to
		the \gls{awbc} (see \fref{tce_arch}). 
	}
\end{assumpB}

\begin{assumpB}
	\rev{The \gls{ste} can be used with any arbitrary terrain geometry given the terrain normal and thus $R_W^{C_i}$. The terrain normal $n_i$ at the contact point $i$ can be
	provided by a height map that is generated via an RGBD sensor.
	}
\end{assumpB}
\begin{algorithm}
	\caption{Terrain Compliance Estimation}\label{tce_alg}
\rev{	\begin{algorithmic}[1]
		\State initialize the buffers ($\mathcal{F}_i^d$ and $\mathcal{P}_i^d$) and $I_i^d$ 
		\For{each iteration $k$} 
		\For{each leg $i$} 
		\If{leg is in contact ($\alpha_i == 1$)}\hfill  \eref{contact_status}
		\For{each direction $d$} 
		\State estimate $\grfest{i}^{d}(k)$ \hfill \eref{grf_est}
		\State estimate $ p_{i}^d(k)$  \hfill \eref{p_est}
		\State estimate $\dot{p}_{i}^d(k)$ \hfill \eref{dotp_est}
		\State transform $\grfest{i}^{d}(k)$ into 	$F_{\mathrm{grf},i}^{C,d}(k)$       \hfill \eref{grf_trans}
		\State transform $p_{i}^{d}(k)$ into 	$p_{i}^{C,d}(k)$           \hfill \eref{p_trans}
		\State transform $\dot{p}_{i}^{d}(k)$ into 	$\dot{p}_{i}^{C,d}(k)$ \hfill \eref{pdot_trans}
		\State update buffers $\mathcal{F}_i^d$ and  $\mathcal{P}_i^d$ \hfill \eref{tce1}-\eref{tce4}
		\State solve for $I_i^d$ \hfill \eref{tce5}
		\EndFor
		\State map the estimated parameters to $\Psi_W$
		\EndIf
		\EndFor 
		\EndFor 
	\end{algorithmic}}
\end{algorithm}
\section{Experimental Setup}
\label{sec:exp_setup}
\subsection{State Estimation}
\label{sec_exp_setup_state}
We implemented our approach on \acrshort{hyq} 
\cite{Semini2011} which is equipped with a variety of sensors. 
Each leg contains two load-cells, one torque sensor, 
and three high-resolution optical encoders.
A tactical-grade \gls{imu} (KVH 1775) is mounted on its trunk.
Of particular importance to this experiment is the Vicon motion capture system (MCS).
It is a multi-camera infrared system capable of measuring the pose of an object with high accuracy.
During experiments, an accurate and non-drifting 
estimate of the position of the feet in the world frame 
is required to calculate the real penetration for the \gls{ste}.
Typically, \acrshort{hyq} works independently of external sensors (\eg MCS or GPS), 
however, soft terrain presents problems for state estimators \cite{Henze2016}.
This was re-affirmed in experiment. 

The current state estimator~\cite{Nobili2017}
relies upon fusion of \gls{imu} and leg odometry data at a high frequency and 
uses lower frequency feedback from cameras or lidars to correct the drift.  
The leg odometry makes the assumption that the ground is rigid. 
On soft terrain, the estimator has difficulties in determining when a foot is in 
contact with the ground (\ie is the foot in the air, or compressing the surface?). 
These errors cause the leg odometry signal to drift jeopardizing the 
estimation. Although, incorporating vision information could be a possibility 
to correct for the drift in the estimation,
improving state estimation on soft 
terrain is an ongoing  area of research and is  out 
of the scope of this paper. 

Despite the drifting problem, we used the current state estimator \cite{Nobili2017} in our \gls{wbc}
because the planner in \fref{fig:blockDiagram} 
has a re-planning feature that
makes our \gls{wbc} robust against a drifting state estimator \cite{Focchi2018}.
However, the \gls{ste} still requires an accurate and non drifting 
estimate of the feet position in the world frame. 
Therefore, to validate the \gls{ste}, we 
used an external MCS that completely eliminates the drift problem.
The MCS measures the pose of a special marker array placed 
on the head of the robot. Then  the position 
of the feet in the world frame was calculated online 
by using the MCS measurement and 
the forward kinematics of the robot. 

\subsection{Terrain Compliance Estimator \gls{ste} Settings}
In this work we used a sliding window of
$m = 250$~samples (or 1 \unit{s} for a control loop 
running at 250 \unit{Hz}).
Despite the general formulation, in this paper we estimate the terrain parameters
only for the direction normal to the terrain, 
and assume that the 
tangential directions are the same. 
\rev{We carried out the simulation and experiment on a horizontal plane. Thus, the rotation matrix $R_W^{C_i}$ is identity.}
Furthermore,
we  did not estimate the damping parameter due to the inherent noise in the feet velocity signals
that would jeopardize the estimation. 
The damping term $D_{\mrm{st}} v_{st}$ in \eref{contactmodel1}
is less dominant in computing the \grfs  
compared to the stiffness term. 
This is because
the feet velocities in the world frame $v_{st}$ are usually orders of magnitude smaller than the penetration
during stance, and
the damping parameter $D_{\mrm{st}}$  is usually orders of magnitude smaller
than the stiffness parameter as shown in \cite{Ding2013}.

\subsection{Tuning of the Low Level Control}
During experiments, we found that the low level torque loop
creates system instabilities when interacting with soft environments.
\rev{
In particular, when we used the same set of (high) torque 
gains in the low level control loop tuned for rigid terrain, we noticed 
joint instabilities when walking over soft terrain.
This is because interacting with soft terrain reduces the stability margins of the system. 
Thus, keeping a high bandwidth in the inner torque loop given the 
reduced stability margin will cause system instability. 
In our previous work~\cite{Focchi2016}, we experimentally validated that increasing 
the torque gain of the inner loop can 
indeed cause system instabilities.}
In fact, this is a well know issue in haptics \cite{Hulin2014}. 
\rev{As a result, reducing the bandwidth by decreasing the torque gains in the inner torque loop was necessary to address these instabilities.}

\rev{
Our control design is a nested architecture consisting of the \gls{wbc} 
and the low level torque control in which,
both control loops contribute to the system stability \cite{Focchi2016,Mosadeghzad2013}.
Over soft terrain, 
the dynamics of the environment also plays a role 
and must be considered in analyzing the stability of the system. 
That said, there is a nontrivial relationship between soft terrain and 
the stability of a nested control loop architecture, 
and a formal and thorough analysis is an ongoing work.}

\section{Results}
\label{sec:results}
In this section, we evaluate the proposed approach on \gls{hyq} in simulation and experiment. 
We compare \textit{three} approaches:
the \gls{swbc} which is the baseline,
the \gls{awbc} which is our proposed \gls{wbc} without the \gls{ste},
and \gls{stance} which incorporates both the \gls{awbc} and \gls{ste}.
We show the extent of improvement given by the \gls{awbc} controller with respect to the \gls{swbc} as well as the 
importance of the \gls{ste} 
during locomotion over multiple terrains with different compliances. 
We set the same parameters and gains throughout the entire simulations and experiments, 
unless mentioned otherwise. 
The results are shown in the accompanying 
video\footnote{Link: \href{https://youtu.be/0BI4581DFjY}{\texttt{https://youtu.be/0BI4581DFjY}}}.

\subsection{Simulations}
\label{sec:sim}
To render soft terrain in simulation, we used the \gls{ode} physics engine \cite{Smith2005}. 
We used \gls{ode} because it is easily integrable with our framework, and
it is numerically fast and stable for stiff and soft contacts \cite{Catto2011}. 
Moreover, \gls{ode} can render soft contacts that emulates physical
parameters (using the SI units  \unit{N/m} and  \unit{Ns/m} for springs and dampers, respectively)
unlike other engines that uses non-physical ones \cite{Erez2015}.
\gls{ode}'s implicit solver uses linear springs and dampers for their soft constraints
which fits perfectly with our contact model \eref{contactmodel1}.
In this way, we have a controlled simulation environment   
where we can emulate any terrain compliance 
by manipulating its stiffness $K_t$ and damping~$D_t$ parameters
similar to our contact model. 
Throughout the simulation, 
we use four types of terrains with the following parameters: 
soft $T_1$ ($K_t=3500$~\unit{N/m}), 
moderate $T_2$ ($K_t = 8000$~\unit{N/m}), 
stiff $T_3$ ($K_t = 10000$~\unit{N/m}), 
and rigid $T_4$ ($K_t =  2\times 10^6$~\unit{N/m})
 all with the  same damping ($D_t = 400$~\unit{Ns/m}).

\subsubsection{Locomotion over Multiple Terrains} 
\label{sec:sim3terrain}
\begin{figure*}[!thb]
	\centering
\includegraphics[width=2\columnwidth]{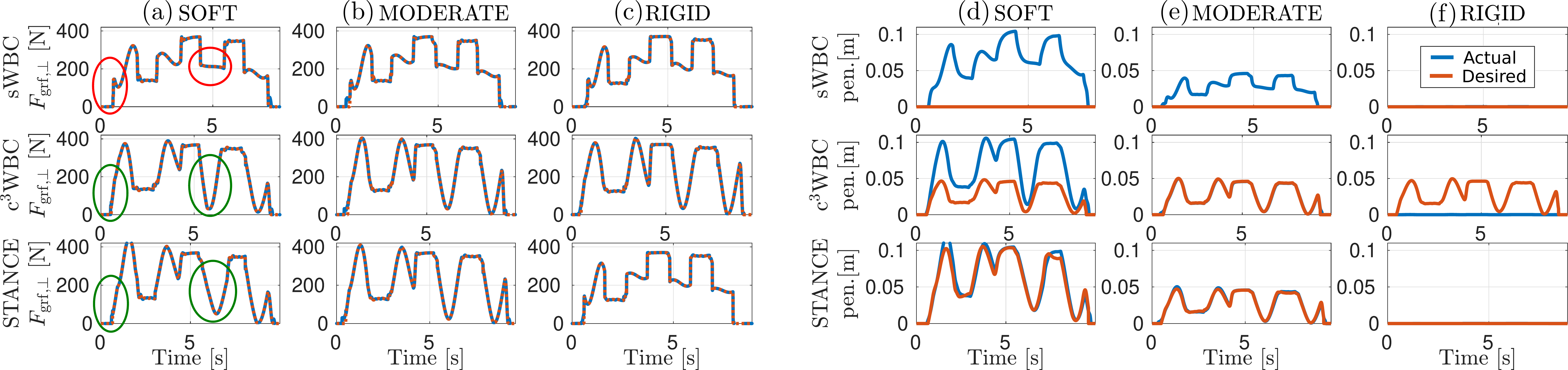}
	\caption{\footnotesize Simulation. Comparison of \gls{swbc}, \gls{awbc}, and \gls{stance} over three type of 
	terrains: 
		soft $T_1$ ($K_t=3500$ \unit{N/m}), 
		moderate $T_2$ ($K_t = 8000$ \unit{N/m}), 
		and rigid $T_4$ ($K_t =  2\times 10^6$ \unit{N/m})
		all with the  same damping ($D_t = 400$ \unit{Ns/m}).
(a)-(c): The actual and desired  contact forces in the normal direction of the \gls{lf} leg for one gait cycle.
(d)-(f): The actual $p$ and desired $\epsilon$ penetration in the normal direction of the \gls{lf} leg for one gait 
cycle.
The red and green ellipses highlight the performance of the three approaches in adapting to soft terrain. 	  
}
	\label{fig:softTerrainSimComparison}
\end{figure*}

\begin{table}[!t]
	\renewcommand{\arraystretch}{1.25}
	 \captionsetup{justification=centering}
	\caption{\footnotesize Mean Absolute Tracking Error (MAE) [N] of the \gls{grfs}
		 in Simulation using \gls{swbc}, \gls{awbc} and \gls{stance}
		 over Multiple Terrains.
		 }
	\label{tab_SIMmeanAbsErrorGRFs}
		\centering
	\begin{tabular}{cccc}
		\hline \hline	
		Terrain & \textbf{\gls{swbc}} & \textbf{\gls{awbc}} & \textbf{\gls{stance}}\\
		\hline
		Soft         	&7.7261  &   7.4419  &  \textbf{6.3547} \\
		Moderate    	& 8.0594 &   \textbf{7.4585}  &  7.9889 \\
		Rigid  	        &\textbf{4.889}   & 6.6523     & 5.128 \\		
		\hline \hline
	\end{tabular}
\end{table}
We evaluate the three approaches with the 
robot walking at $0.05$ \unit{m/s} over the terrains:
 $T_1$ (soft), $T_2$ (moderate), and $T_4$ (rigid).
We provided the \gls{awbc} with the terrain parameters of the moderate terrain~$T_2$
for all the three simulations. 
We do that in order to test the performance of \gls{awbc} if given
the real terrain parameters (in case of $T_2$)
or inaccurate parameters (in case of  $T_1$ and $T_4$).
In this simulation, 
we compare the actual values of $ \grfp{\perp}$ against the optimal values 
$\grfp{\perp}^*$(solution of the \gls{wbopt}) as well as the actual penetration $p$ against the 
desired penetration~$\epsilon$ of the \gls{lf} leg.
We have omitted the other three feet for space as all four legs have the same performance.
The results are shown in \fref{fig:softTerrainSimComparison}.
The \gls{mae} of the \grfs in these simulations are presented in 
\tref{tab_SIMmeanAbsErrorGRFs}.
The \gls{mae} of the \grfs is defined as: $\text{MAE} = \frac{1}{T} \int_{0}^{T} \vert
	F_{\mrm{grf}} - F_{\mrm{grf}}^* \vert dt $.

\fref{fig:softTerrainSimComparison}a
captures the effect of the three approaches
on the \grfs over soft terrain. 
We can see that a \gls{wbc} based on a rigid contact assumption (\gls{swbc}) assumes that it can achieve
an infinite bandwidth from the terrain and thus supplying an instantaneous change in the \gls{grfs} as highlighted by 
the red ellipses 
in \fref{fig:softTerrainSimComparison}a. 
On the other hand, \gls{stance} and \gls{awbc} were both 
capable of attenuating this effect as highlighted by the green ellipses.
For the reasons explained earlier in this paper and in \cite{Grandia2019},
instantaneous changes in the \gls{grfs} are undesirable over soft terrain. 
This resulted in an improvement in the tracking of the \grfs  
in \gls{stance} and \gls{awbc} compared to \gls{swbc} as shown
 in \tref{tab_SIMmeanAbsErrorGRFs}. 
Moreover, by comparing \gls{awbc} and \gls{stance} over soft terrain $T_1$, 
we can see that the shape of the \grfs did not differ. 
However, the tracking of the \grfs in \gls{stance} is better than the 
\gls{awbc}.
This shows that suppling the \gls{awbc} with the incorrect values of the terrain 
parameters 
deteriorates the \grfs tracking performance.
\fref{fig:softTerrainSimComparison}b shows the \grfs on a moderate terrain.
Since the \gls{awbc} is provided with the exact terrain parameters of $T_2$, 
we can perceive the \gls{awbc} as \gls{stance} with a perfect \gls{ste} on moderate terrain. 
As a result, \tref{tab_SIMmeanAbsErrorGRFs} 
shows that in this set of simulations, 
\gls{awbc} outperformed \gls{stance}
in
the \grfs tracking.
 This shows  that a  more accurate \gls{ste}
 can result in a better \grfs tracking.
Additionally, 
\fref{fig:softTerrainSimComparison}c shows the \grfs on rigid terrain.
We can see that  the \gls{swbc} resulted in a typical (desired) shape of the \grfs 
for a crawl motion in rigid terrain \cite{Focchi2018}. 
\gls{stance} showed a shape of the \grfs similar to the \gls{swbc} which
is expected since the \gls{ste} provided \gls{stance} with  parameters similar to the rigid terrain.
However, for \gls{awbc}, the \grfs shape did not change compared to the other three terrains. 
As shown in \tref{tab_SIMmeanAbsErrorGRFs}, the best tracking to the \grfs was by the 
\gls{swbc}, which was expected since the \gls{swbc} was designed for rigid terrain. 
However, \gls{swbc} was only slightly better than \gls{stance} due to small estimation errors from the \gls{ste}.

\fref{fig:softTerrainSimComparison}a-c show the superiority of \gls{stance} compared
to \gls{swbc} and \gls{awbc}. \gls{stance} adapted to the three terrains
by estimating their parameters and supplying them to the \gls{wbc}.
This resulted in changing the shape of the \grfs accordingly 
that improved the tracking of the  \gls{grfs}.  
Unlike \gls{stance}, the \gls{swbc} and the \gls{awbc} both are contact consistent for only \textit{one} type of terrain
which resulted in a deterioration of the \grfs tracking over the other types of terrains.
The advantages of \gls{stance} compared to \gls{swbc} and \gls{awbc} are also shown in 
\fref{fig:softTerrainSimComparison}d-f. 
Since the \gls{swbc} is always assuming a rigid contact, 
the penetration $\epsilon$
was always zero throughout the three terrains. 
Similarly, since the \gls{awbc} alone is aware only of one type of terrain, 
it is always assuming the same contact model, 
in which the desired penetration $\epsilon$ was similar throughout the three terrains.  
\gls{stance}, however, was capable of predicting the 
penetration correctly for all the three terrains. 

In general,
even if the contact model is for soft contacts, \gls{stance} was capable of correctly
predicting the penetration of the robot even in rigid terrain (zero penetration). 
This resulted in \gls{stance} adapting to rigid, soft and moderate terrains by means
of adapting the \grfs and correctly predicting the penetration. 

\subsubsection{Longitudinal Transition Between Multiple Terrains}
\label{sec:simCrossingTerrain}
\begin{figure}[!t]
	\centering
	\includegraphics[width=\columnwidth]{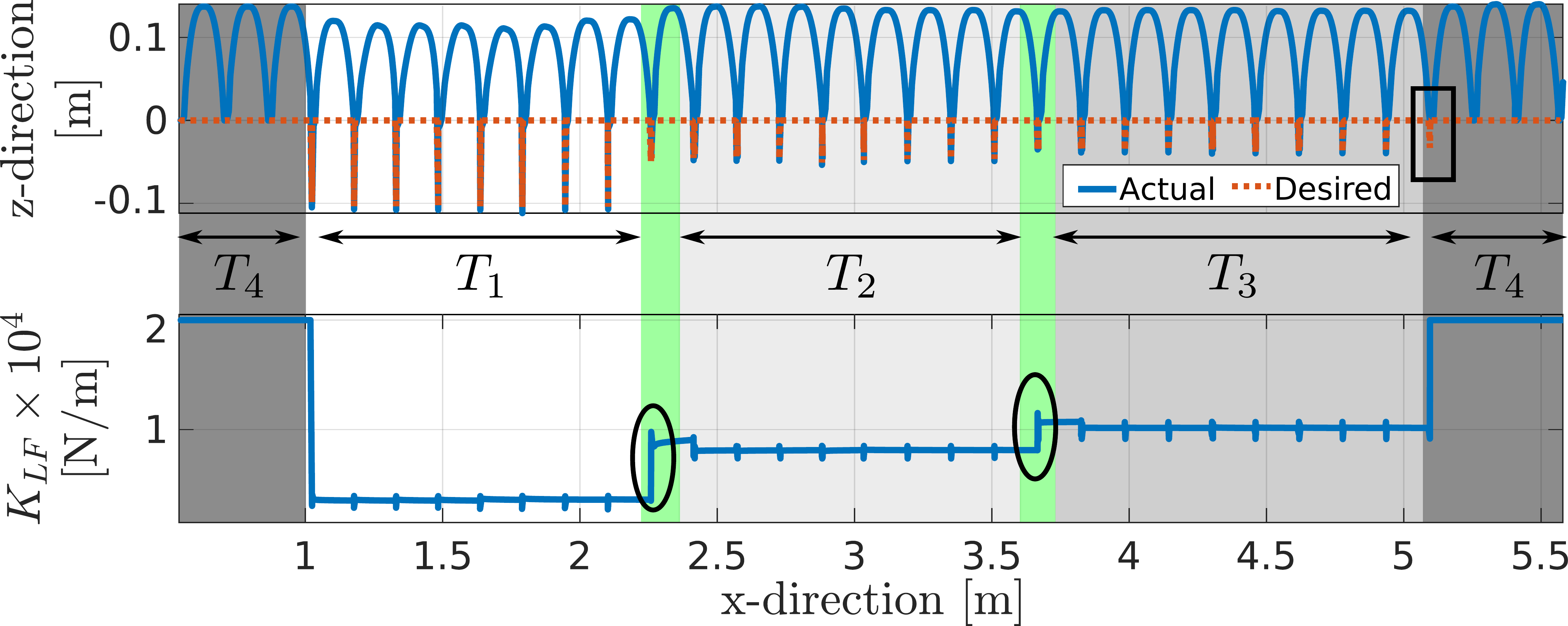}
	\caption{\footnotesize Simulation. Traversing multiple terrains of different compliances ($T_4$, $T_1$, $T_2$, $T_3$, 
	$T_4$). 
		Top: Tracking of the desired terrain penetration of the \gls{lf} leg  in the xz-plane. 
		Bottom: Estimated terrain stiffness of the \gls{lf} leg. For readability purposes we only plot estimated values less than $2\times10^4$.
		The green shaded areas highlight the overlap 
		between terrains that results in higher estimated stiffness (black ellipses).}
	\label{fig:simCrossingTerrain}
\end{figure}
We show the adaptation of \gls{stance} when walking and transitioning between 
multiple terrains. 
We test the accuracy of the \gls{ste} and 
the effect of closing the loop of the \gls{awbc} with the \gls{ste} 
on the feet trajectories and terrain penetration. 
In this simulation, \gls{hyq} is traversing five different terrains, starting and
ending with a rigid terrain: $T_4$, $T_1$, $T_2$, $T_2$, $T_4$.
The results are presented in \fref{fig:simCrossingTerrain}. 
The top plot presents the actual foot position against the desired penetration
$\epsilon$
of the \gls{lf} leg in the xz-plane of the world frame. 
The origin of the z-direction (normal direction) 
is the uncompressed terrain height. 
Thus, trajectories below
zero represent the penetration of the \gls{lf} leg. 
The bottom plot shows the history of the
estimated terrain stiffness of the  \gls{ste} of the
\gls{lf} leg.  
Table \ref{tab_stesim} reports the mean, standard deviation, 
and percentage error\footnote{ The percentage error is defined as:
	\% Error $= \vert \frac{\text{Estimate}-\text{Actual}}{\text{Actual}} \vert \times100$
	}
of the estimated terrain stiffness of the \gls{lf} leg 
against the ground truth value set in \gls{ode}.
The table shows that the \gls{ste} had an estimation accuracy below $2$\% for the soft terrains $T_1$, $T_2$ and $T_3$. 
However, the estimation accuracy of the rigid terrain was lower than that of the soft terrains. 
This is expected since on a rigid terrain, the penetrations are (almost) zero. 
Thus, a small inaccurate penetration estimation due to any model errors could 
result in a lower estimation accuracy. 
Apart from the rigid case, the standard deviation 
is always below~6$\%$ of the ground truth value. 
\fref{fig:simCrossingTerrain} shows that \gls{stance}
is always \gls{c3},  
the actual foot position is always consistent with the 
desired penetration
during stance. 
We can see that, when \gls{hyq} is standing over rigid terrain, 
both the actual foot position and desired penetration are zero. 
As \gls{hyq} walks, over the soft terrains, the penetration is highest in the softest terrain
and smallest in the stiffest terrain. 

In the simulation environment, we overlapped the 
terrains to prevent the feet from getting stuck between them. 
This overlap created a transition (highlighted in green in the figure) which resulted in a stiffer 
terrain. 
The overlap was captured by the \gls{ste} and resulted in a slight increase in the estimated 
parameters as highlighted by the two ellipses in the lower plot.
We also noticed a lag in estimation, 
due to a filtering effect,
since the \gls{ste} is using the most recent $m$-samples.
As highlighted by the black box in \fref{fig:simCrossingTerrain}, 
\gls{hyq} was on rigid terrain (actual penetration is zero) while \gls{stance}  still perceived
it as being on $T_3$ (desired penetration is non-zero).

\begin{table}
	\centering
	\caption{\footnotesize Mean $\mu$ [N/m], Standard Deviation $\sigma$ [N/m], and Percentage Error  of the 
		Estimated Terrain Stiffness 
		of the \gls{lf} Leg in Simulation.}
	\label{tab_stesim}
	\renewcommand{\arraystretch}{1.25}
	\vspace{-0.2cm}
	\begin{tabular}{c ccc}
		\hline \hline
		Terrain & Actual Stiffness & Mean $\mu$ $\pm$ STD $\sigma$ & \% Error\\
		\hline
		$T_1$ & 3500 & 3530 $\pm$ 200 & 0.9\% \\ 
		$T_2$ & 8000 & 8110 $\pm$ 400 & 1.4\% \\
		$T_3$ & 10000 &  10110 $\pm$ 400~& 1.1\%  \\
        $T_4$ & 2000000 & 2240000 $\pm$ 740000 & 12\%  \\
		\hline \hline
	\end{tabular}
\end{table}

\subsubsection{Aggressive trunk maneuvers} 
\label{sec:simAggressiveTrunk}
\begin{figure}[!t]
	\centering
	\includegraphics[width=\columnwidth]{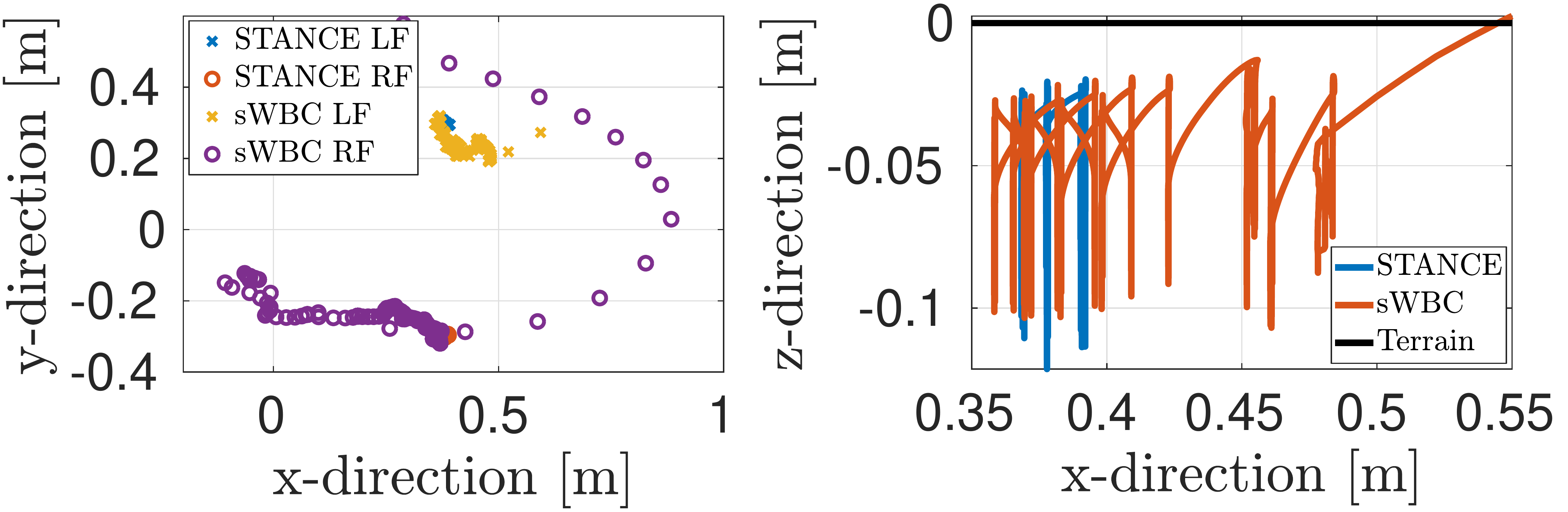}
	\caption{\footnotesize Simulation. 
		Comparing \gls{swbc} and \gls{stance}
		under aggressive trunk maneuvers. 
		Left: Top view of the front feet (\gls{rf} and \gls{lf}) positions. 
		Right: Side view of the  \gls{lf}  position.
			}
	\label{fig:simAggressiveTrunkTraj}
\end{figure}
We tested \gls{swbc} and \gls{stance}  under aggressive trunk maneuvers by
commanding  desired sinusoidal trajectories at the robot's height ($0.05$ \unit{m} amplitude and 
$1.8$ \unit{Hz} frequency) and at roll orientation ($0.5$ \unit{rad} amplitude and $1.5$ \unit{Hz} frequency) 
over the soft terrain $T_1$.
The results are shown in \fref{fig:simAggressiveTrunkTraj}. 
The left plot shows a top view of the actual front feet (\gls{lf} and \gls{rf}) positions in the world frame.
The right plot shows a side view of the actual  \gls{lf} foot position in the world frame. 
We notice that 
the feet of \gls{hyq} are always in contact with the terrain
in \gls{stance}
which is expected since \gls{stance} is \gls{c3}.
Unlike \gls{stance},
the feet did not remain in contact with the terrain
in the \gls{swbc}. 
This is clearly seen in \fref{fig:simAggressiveTrunkTraj} where
\gls{hyq} lost  contact multiple times.
 This resulted
in the robot falling over in the \gls{swbc} case as shown in the video.

\subsubsection{Speed Test}
\label{sec:simSpeedTest}
\begin{figure}[!t]
	\centering
	\includegraphics[width=0.9\columnwidth]{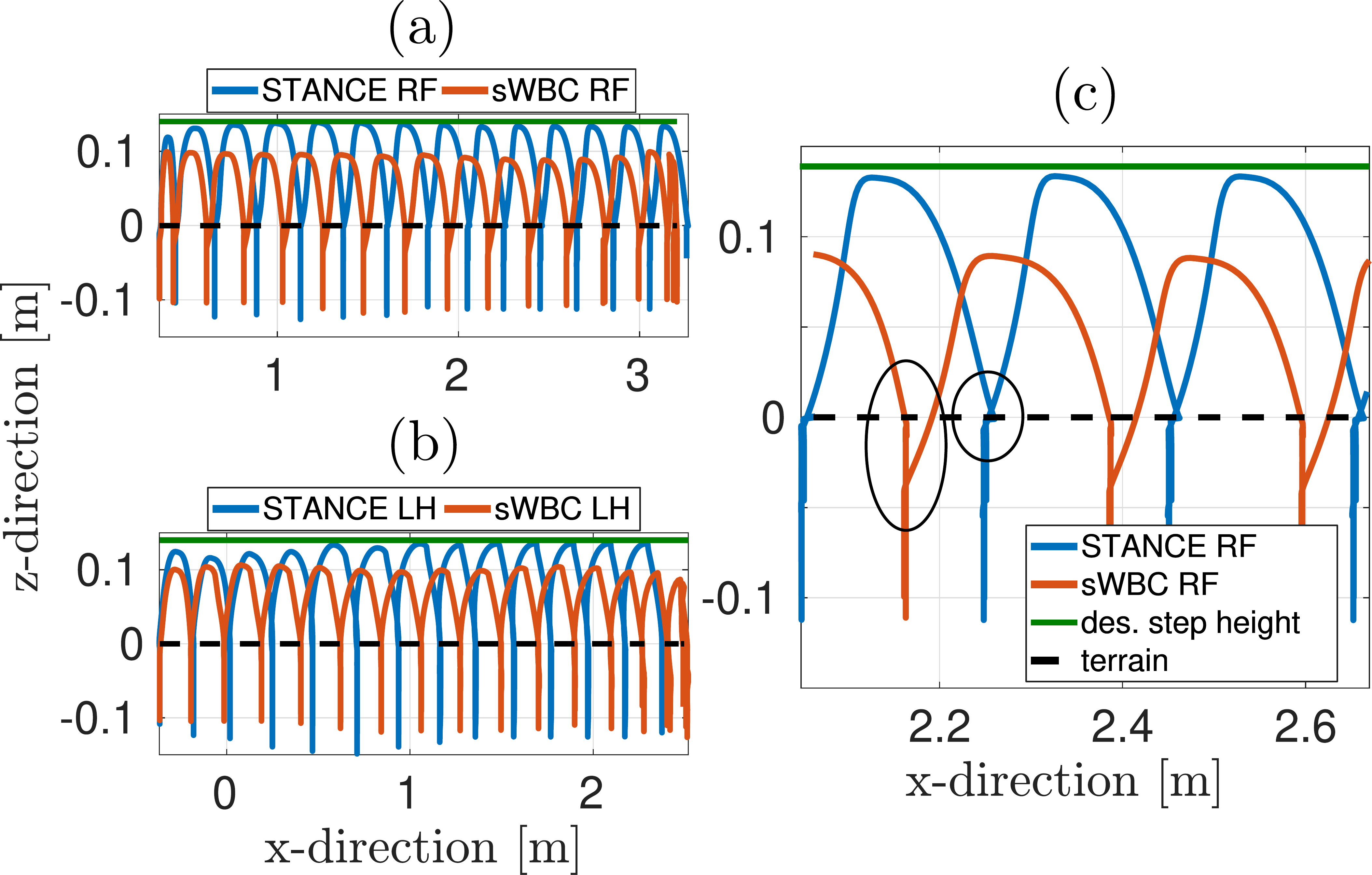}
	\caption{\footnotesize Simulation. Speed test. Increasing the desired forward velocity from $0.05$ to $0.3$ \unit{m/s}.
	Left: Side view of the \gls{rf} (a) and \gls{lh} (b) positions. 
	Right:~(c)~Closeup section of the top left plot. 
	The green lines are the desired step height. 
	The black dashed lines are the terrain height.
}
	\label{fig:speedTest}
\end{figure}
We carried out a simulation 
where \gls{hyq} walks over soft terrain $T_1$,
starting with a forward 
velocity of $0.05$  \unit{m/s} until it reaches $0.3$  \unit{m/s} 
with an acceleration of~$0.005$~\unit{m/s$^2$}.
In this simulation, 
we compare \gls{stance} against the \gls{swbc}. 
\fref{fig:speedTest}a and \fref{fig:speedTest}b
show the actual
trajectories
of the  \gls{rf} and \gls{lh} legs
in the world frame, respectively. 
\fref{fig:speedTest}c shows a closeup  section of
the
\gls{rf} leg's trajectory. 
The simulation shows that \gls{stance} was \gls{c3} over the entire simulation
while the \gls{swbc} was not.

In particular, \gls{stance} was able to remain in contact with the terrain that allowed \gls{hyq} to  start the swing phase directly from the terrain height. 
Unlike \gls{stance},   the \gls{swbc} is not terrain aware
and did not remain \gls{c3} which resulted in
	starting the swing trajectory while still being inside the deformed terrain. 
This is highlighted by the two ellipses in the right plot.
Additionally, the compliance contact consistency property of \gls{stance} enabled the robot to 
maintain the desired step clearance
(i.e., achieving the desired step height of $0.14$ \unit{cm})
compared to \gls{swbc}. 
Most importantly, 
as shown in the accompanying video, 
the \gls{swbc} failed to complete the simulation and could not achieve the final desired forward velocity;  It fell at a speed of $0.21$~\unit{m/s}.
Note that  both approaches could reach higher velocities
with a more dynamic gait (trot). However, 
this simulation is not focusing on analyzing the
maximum speed that the two approaches can reach
but rather the differences between these approaches
at a higher crawl speeds. 

\subsubsection{Power Test}
\label{sec:powertest}
\begin{figure}[t!]
	\centering
	\includegraphics[width=\columnwidth]{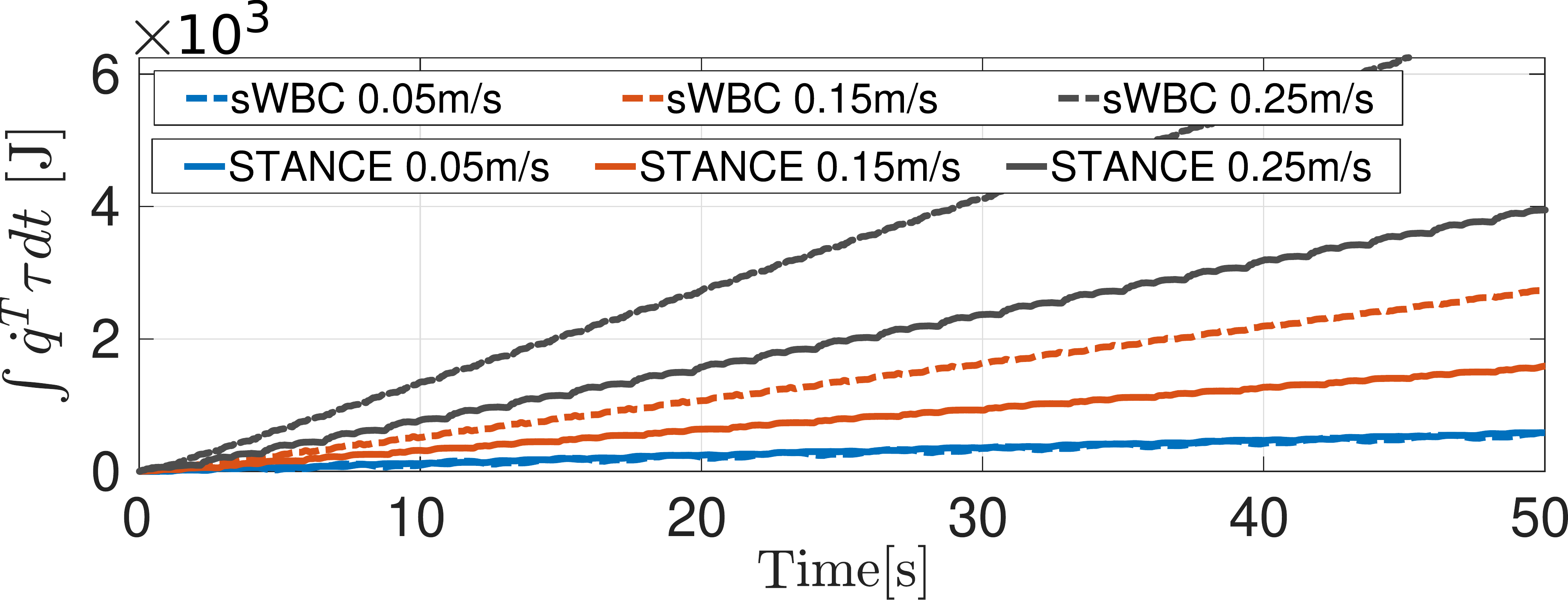}
	\caption{\footnotesize Simulation. Power consumption comparison between
		\gls{swbc} and \gls{stance} with different forward velocities (0.05~\unit{m/s}, 0.15~\unit{m/s} and 	 0.25~\unit{m/s}). 
	}
	\label{fig:powerConsumption}
\end{figure}
In this test, 
we compare the power consumption using
\gls{stance} and \gls{swbc}
on
\gls{hyq} during walking over the  soft terrain $T_1$
at different forward velocities ($0.05$~\unit{m/s}, $0.15$ \unit{m/s} and $0.25$ \unit{m/s}).
\fref{fig:powerConsumption} presents the energy plots of \gls{stance} and \gls{swbc}. 
The plot shows that \gls{stance}  requires less power than the \gls{swbc} 
because it knows how the terrain will deform. 
\gls{stance} exploits the terrain interaction
to achieve the motion. 
The difference in consumed energy is 
negligible at $0.05$ \unit{m/s} but becomes significant
at higher speeds.

\subsection{Experiment}
\label{sec:exp}
We validated the simulation presented in \sref{sec:sim} on the real platform. 
We analyzed \gls{swbc},  \gls{awbc} (with fixed terrain parameters) and \gls{stance}
as well as the performance of the \gls{ste} module itself. 

A foam block of 
$160$~\unit{cm}~$\times~120$~\unit{cm}~$\times~20$~\unit{cm} 
was selected as a soft terrain
for these experiments.
To obtain a ground truth of the foam stiffness, 
we carried out indentation tests on a $50$ \unit{cm}$^3$ sample  of 
the foam with a stress-strain machine that covers the 
range of penetration of interest for our robot (below $0.15$ \unit{cm}). 
The indentation test  showed a softening behavior of the foam with an average stiffness 
of $2400$ \unit{N/m}.
The \gls{mae} of the \grfs of the upcoming experiments are shown in
 \tref{tab_EXPmeanAbsErrorGRFs}.

\subsubsection{Locomotion over Soft Terrain}
\label{sec:expsSoftTerrain}
\begin{figure}[!t]
	\centering
	\includegraphics[width=\columnwidth]{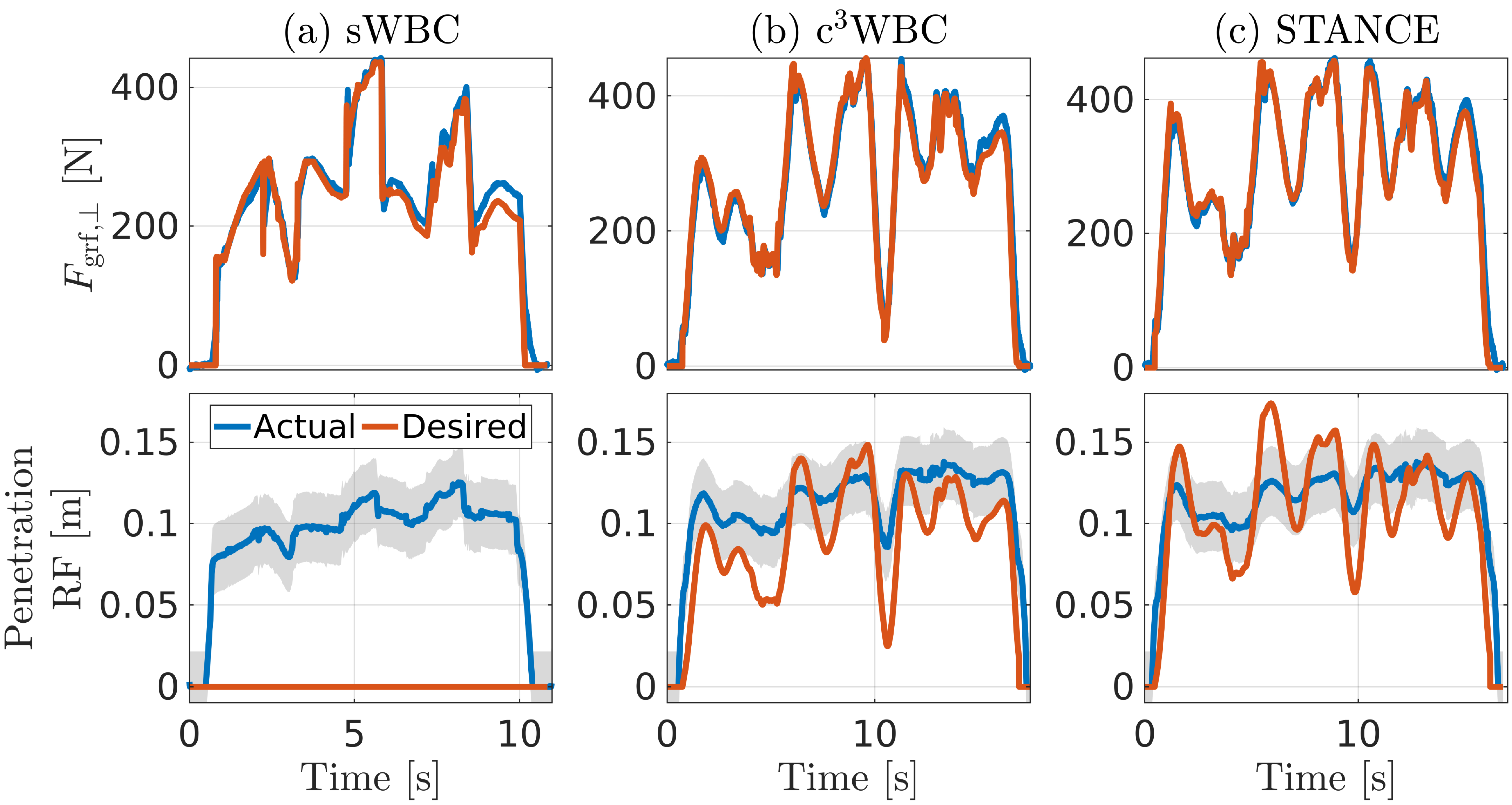}
	\caption{\footnotesize Experiment. 
		Comparing \gls{swbc}, \gls{awbc}   
		and \gls{stance} over a soft foam  
		 block ($K_t = 2400$ \unit{N/m}).  
Top: Tracking of the \grfs of the \gls{rf} leg. 
		Bottom: Tracking of the foot penetration. 
		The gray shaded areas represent the uncertainty of the measurements.
	}
	\label{fig:expsSoftTerrain}
\end{figure}
\begin{table}[t]
	\centering
	\caption{\footnotesize Mean Absolute Tracking Error (MAE) [N] of the \gls{grfs}
		using \gls{swbc}, \gls{awbc} and \gls{stance}
		under Different Sets of Experiments.}
	\label{tab_EXPmeanAbsErrorGRFs}
	\renewcommand{\arraystretch}{1.25}
	\begin{tabular}{lccc}
		\hline \hline	
		Description & \textbf{\gls{swbc}} & \textbf{\gls{awbc}} & \textbf{\gls{stance}}\\
		\hline
		Soft Terrain (Sec. \ref{sec:expsSoftTerrain})    &	73.9042		& 68.5581 	 & 
		\textbf{61.8207	}	\\
		Longitudinal Trans. (Sec. \ref{sec:expCrossingTerrain})     &	 70.5276	&  64.2636 	 & 	
		\textbf{60.6285} 		\\
		Lateral Trans. (Sec. \ref{exp_lat})  		&	73.0766		& 	- 	 & 	\textbf{53.0107}		
		\\
		\hline \hline
	\end{tabular}
\end{table}
\begin{table}
	\centering
	\caption{\footnotesize Mean $\mu$ [N/m], Standard Deviation $\sigma$ [N/m], and Percentage Error of the 
		Estimated Terrain Stiffness of the Four Legs in Experiment over Soft Terrain ($2400$~\unit{N/m}).}
	\label{tab_ste_exp_soft}
	\renewcommand{\arraystretch}{1.25}
	\vspace{-0.2cm}
	\begin{tabular}{c ccc}
		\hline \hline
		Leg & Mean $\mu$ $\pm$ STD $\sigma$ & \% Error\\
		\hline
		LF & 2186 $\pm$ 166 & 9\% \\ 
		RF & 2731 $\pm$ 173 &14\% \\
		LH & 2368 $\pm$ 317~& 1\%  \\
		RF & 2078 $\pm$ 331 & 13\%  \\
		\hline \hline
	\end{tabular}
\end{table}

In this experiment, \gls{hyq} is walking over the foam with a forward velocity of $0.07$~\unit{m/s}
using the three approaches. 
The results  are presented in \fref{fig:expsSoftTerrain}
that shows the actual and desired $\grfp{\perp}$ and penetration of the \gls{rf} leg.
The shaded gray area in the lower plots of \fref{fig:expsSoftTerrain} represents the uncertainty in the estimation 
of the foot position 
(see \sref{sec_exp_setup_state}).
In these experiments, all three approaches performed well; none of them failed. 
However, the shape of \grfs 
were different within the three approaches. 
As in \sref{sec:sim3terrain}, 
since \gls{swbc} is rigid contact consistent, the desired \grfs were designed for rigid contacts. 
Unlike \gls{swbc}, \gls{stance} is \gls{c3}, which was capable of changing the shape of the \gls{grfs}. 
This is highlighted in \tref{tab_EXPmeanAbsErrorGRFs} in which \gls{stance} 
outperformed \gls{swbc} in the tracking of the \gls{grfs}. 

In simulation,
 when we provided the \gls{awbc} with the true value
of the  stiffness, the
\gls{mae} of the \gls{grfs} was better. 
However, in this experiment, providing the value obtained
from the indentation tests to the \gls{awbc} resulted in a worse \grfs \gls{mae}.
This outperformance of \gls{stance} compared to the \gls{awbc}
in this experiment could be because of the \gls{ste}. 
To clarify, 
the actual terrain compliances are not constant, 
but since the \gls{ste} is online, it is able to capture these changes in the 
terrain compliances as well as model errors. 
As shown in the accompanying video, 
\gls{stance} had a smoother transition during crawling compared to \gls{swbc}. 
We found the robot transitioning from swing to stance more aggressively in \gls{swbc} 
than \gls{stance}. Such smooth behavior was also noticed in \cite{Grandia2019}.

\tref{tab_ste_exp_soft} shows the mean, standard deviation, 
and percentage error 
of the estimated terrain stiffness of all the four legs 
against the ground truth value ($2400$~\unit{N/m}) obtained from the indentation tests.
The table shows that the accuracy of the \gls{ste} in simulation is better than in experiments. 
This is expected since in simulation, the \gls{ste} has a perfect knowledge of the feet penetration. 
However, the accuracy of our \gls{ste} is better compared to \cite{Bosworth2016} in which the percentage error
exceeded 50\% (the actual stiffness was more than double that of the estimated one in \cite{Bosworth2016}).

\subsubsection{Longitudinal Transition Between Multiple Terrains}
\label{sec:expCrossingTerrain}
\begin{figure}[!t]
	\centering
	\includegraphics[width=1.0\columnwidth]{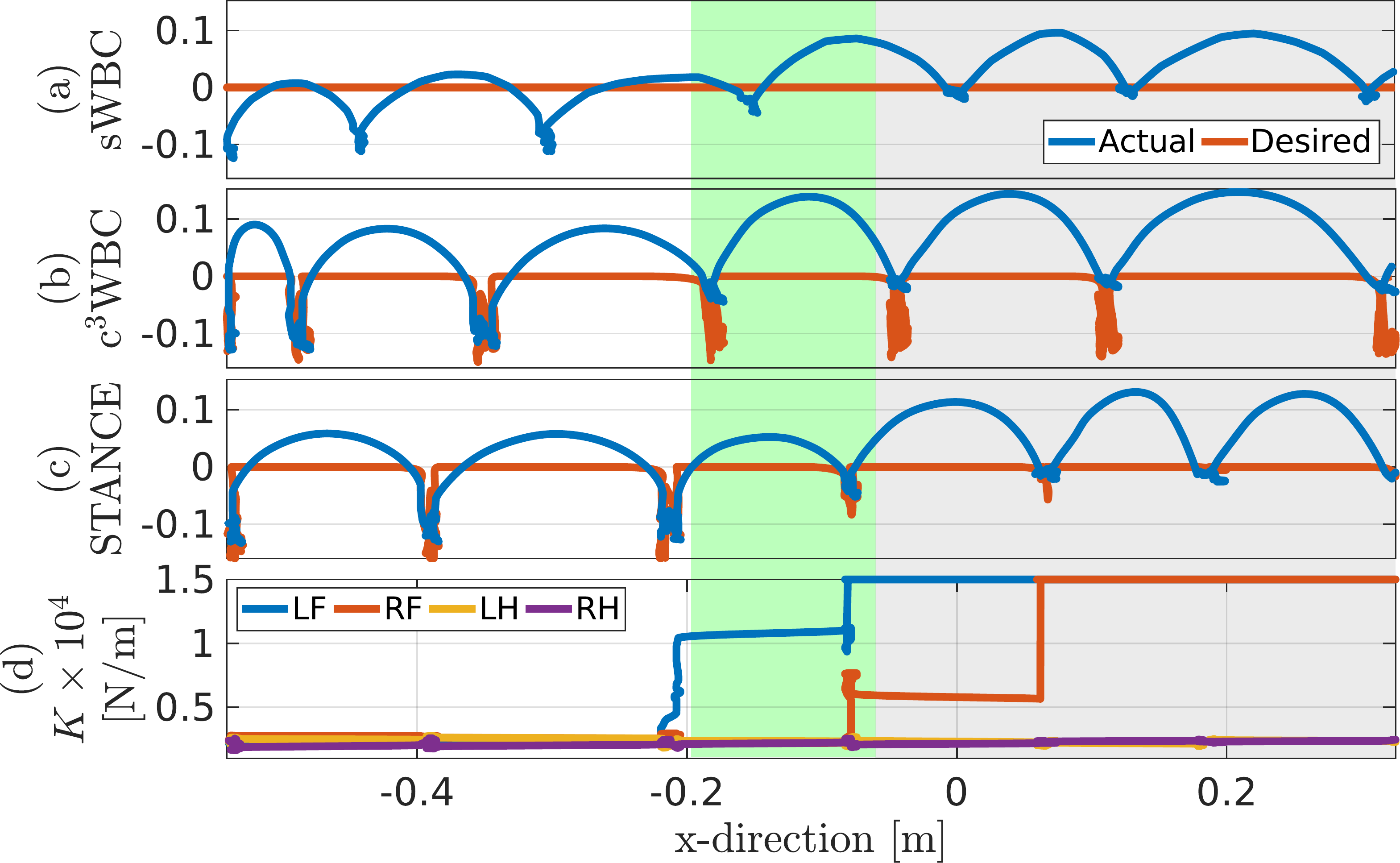}
	\caption{\footnotesize Experiment.
		Longitudinal transition from soft to rigid terrain. 
		The first three plots show the tracking of the 
		desired foot penetration of \gls{rf} leg using the three approaches
		(the \gls{swbc}, \gls{awbc} with fixed terrain stiffness  and \gls{stance}.
		The fourth plot shows the stiffness estimated by the \gls{ste} for the four legs.}
	\label{fig:longitudinalArrangement}
\end{figure}
Similar to \sref{sec:simCrossingTerrain},  we compare the three approaches while
transitioning between the foam block and a rigid pallet. 
We added a pad between between the two terrains to avoid the 
foot getting stuck (see \fref{fig:photos}a).
\fref{fig:longitudinalArrangement}a-c
show the actual position and the desired penetration of the \gls{rf} leg 
in the  xz-plane for the three approaches. 
\fref{fig:longitudinalArrangement}d shows the estimated terrain stiffness of the \gls{ste} for all four  feet.

From \fref{fig:longitudinalArrangement}a-b we see that both \gls{swbc} and \gls{awbc} 
did not adapt to terrain changes. 
Since both controllers are designed for a specific constant terrain,
the desired penetration did not change from soft to rigid. 
In the \gls{swbc}, there is no tracking of the penetration,
and in the \gls{awbc}, the tracking of the penetration is good only when the leg is on the foam 
where the stiffness is consistent to the one used in the controller.
On the other hand, as shown in \fref{fig:longitudinalArrangement}c-d, \gls{stance} changes its parameters when 
facing a 
different terrain; it was capable of adapting its desired penetration to the type of terrain. 
In fact, the desired penetration was non-zero on soft terrain and was almost zero on rigid terrain. 
This again resulted in \gls{stance} achieving the 
best \grfs tracking as shown in  \tref{tab_EXPmeanAbsErrorGRFs}.

\fref{fig:longitudinalArrangement}d shows the importance of having a \gls{ste} for each leg. 
The estimated terrain parameters are different between the legs where 
the hind legs are on the foam while the rigid ones transitioning from foam to rigid. 
The figure also shows that the \gls{lf} leg walked over the rigid terrain before the \gls{rf} 
and that the \gls{ste} captures the 
intermediate stiffness estimation due to the rubber pad (see video).

\subsubsection{Lateral Transition Between Multiple Terrains}
\label{exp_lat}

Unlike the previous experiment, we set the foam and the pallet laterally as shown in \fref{fig:photos}c 
and in the accompanying video.
This is a more  challenging scenario for stability reasons. 
In particular, the robot must extend its leg further in the soft terrain 
maintain the trunk's balance. 
Consequently, since the width of \gls{hyq}'s torso is smaller than its length, 
the \gls{zmp} is more likely to get out of the support polygon.
The \gls{grfs} \gls{mae} in \tref{tab_EXPmeanAbsErrorGRFs} show that \gls{stance} can outperform \gls{swbc} during 
both
longitudinal and lateral transitions.

\subsubsection{External Disturbances over Soft Terrain}
\begin{figure}[!t]
	\centering
	\includegraphics[width=0.95\columnwidth]{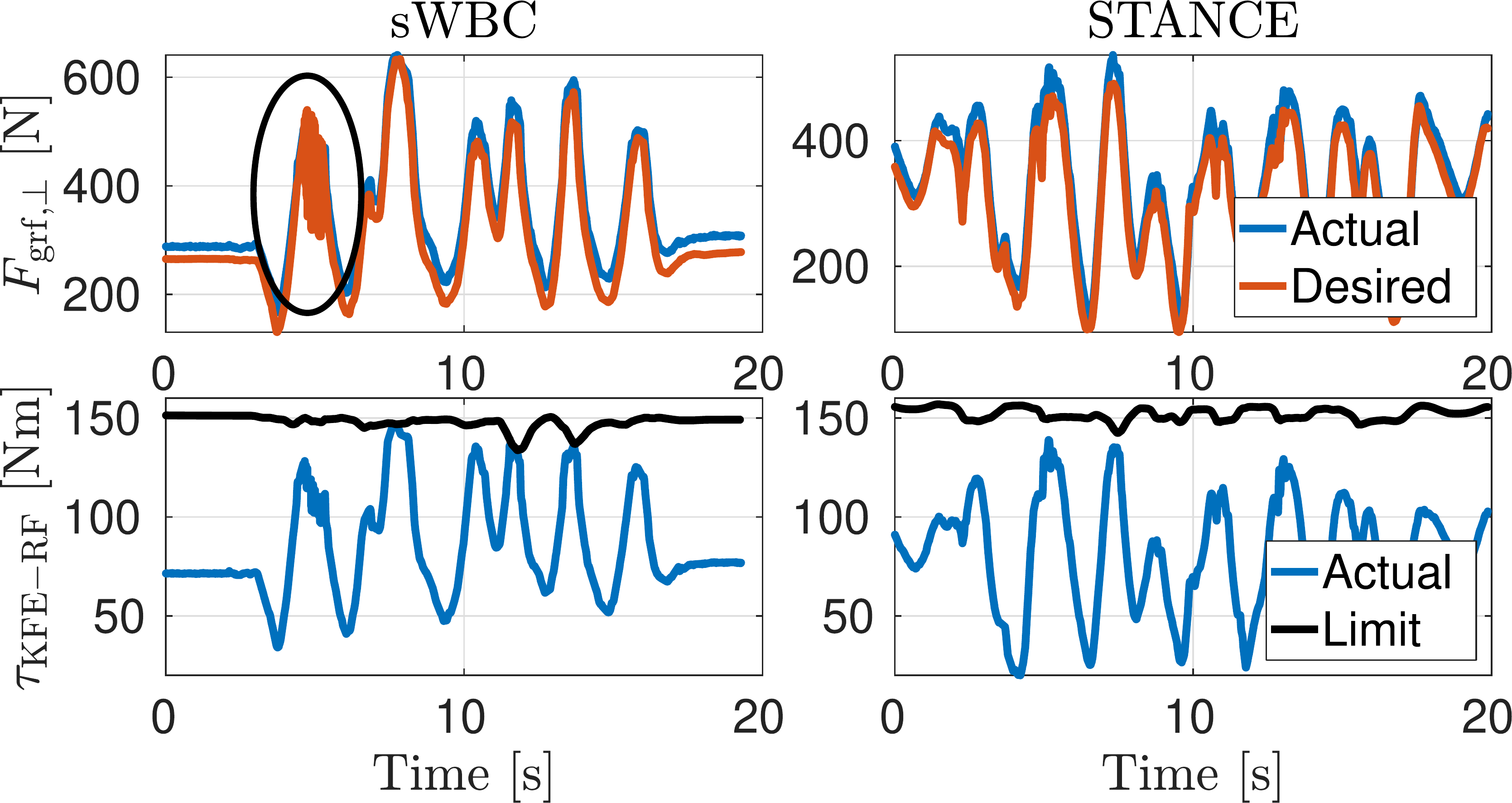}
	\caption{\footnotesize Experiment. The \gls{swbc} and \gls{stance}
		under disturbances over soft terrain. 
		Top: The actual and desired $\grfp{\perp}$ in \gls{swbc} and \gls{stance}, respectively.
		Bottom: The actual torque and torque limits of the Knee Flexion-Extension (KFE) joint of the \gls{rf} 
		leg
		in \gls{swbc} and \gls{stance}, respectively.
	}
	\label{fig_exp_dist}
\end{figure}
In this experiment, we test the \gls{swbc} and \gls{stance} 
when the user applies a disturbance on \gls{hyq}.
The results are shown in \fref{fig_exp_dist}. 
The top plots show the actual and desired $\grfp{\perp}$ in \gls{swbc} and \gls{stance}, respectively.
The bottom plots show the actual torque and torque limits of the \gls{kfe} joint of the \gls{rf} leg
in \gls{swbc} and \gls{stance}, respectively.
In the accompanying video, we can qualitatively see that with \gls{stance}, the feet of \gls{hyq} keep moving 
to remain \gls{c3} with the terrain. 
On the other hand, the \gls{swbc} kept its feet stationary. 
This behavior was also reported by \cite{Henze2016}. 

Most importantly,   we noticed that \gls{hyq}
reaches the torque limits in the \gls{swbc}
as shown  in \fref{fig_exp_dist}.
 However, in \gls{stance}, since the robot was
constantly moving its feet, hence redistributing its forces, 
the torque limits were not reached. 
This behavior was also reflected on the \grfs 
in which, the \grfs were resonating in the \gls{swbc} as highlighted by 
the ellipse in \fref{fig_exp_dist}.

\subsubsection{\gls{ste}'s Performance over Multiple Terrains}

\begin{table}[t!]
	\centering
	\caption{\footnotesize Mean $\mu$  \unit{[N/m]} and Standard Deviation $\sigma$
		[\unit{N/m}]
		of the Estimated Terrain Stiffness of the Four Legs in Experiments (see \fref{fig:photos}b).}
	\label{tab_ste_exp}
	\renewcommand{\arraystretch}{1.25}
	\vspace{-0.2cm}
	\begin{tabular}{ccc}		
		\hline \hline	
Leg & Mean $\mu$ $\pm$ STD $\sigma$ \\
		\hline
	LF &	448400 $\pm$ 165100    \\     
RF 		&55200 $\pm$ 48400\\
	LH 	&2645000 $\pm$ 336000\\
		RH  &  1393000 $\pm$ 442000 \\
		\hline \hline
	\end{tabular}
\end{table}

We analyze the performance of the \gls{ste} on \gls{hyq} over multiple terrains with various softnesses. 
The softness of the four used terrains are shown in \fref{fig:photos}b. 
The estimated stiffness (mean and standard deviation) under each leg is shown in \tref{tab_ste_exp}.
As shown in the table, the robot can  differentiate between the types of 
terrain.
Although we did not measure the true stiffness value of these terrains,
we can observe their softness 
in the video and \fref{fig:photos}b
and compare it to the values in  \tref{tab_ste_exp}.

\subsection{\rev{Computational Analysis}}
\rev{
\gls{stance} is running online which means that we can
estimate the terrain compliance (using the TCE) continuously while walking, 
and run the entire framework without breaking real-time requirements. 
We validated the first argument by showing that indeed the \gls{ste} 
can continuously estimate the terrain compliance. 
Hereafter, we validate the second argument by analyzing the computational complexity of 
\gls{stance} and compare it against the \gls{swbc}. 
Since our \gls{wbc} framework is running at $250~\unit{Hz}$, 
it is essential that the computation does not exceed the $4~\unit{ms}$ time frame.  
}
\rev{
Hence, 
we conducted a simulation in which we calculated the time 
taken to process the entire framework 
without the lower level control 
(ie., the state estimator, the planner and the \gls{wbc})
that is running on a different real-time thread at $1~\unit{kHz}$. 
We compared the computation time 
on an Intel Core i$7$ quad core CPU
in the case of \gls{stance} 
(the \gls{awbc} and the \gls{ste}) and the \gls{swbc}.
We used the same parameters and gains as in 
\sref{sec:sim3terrain}.
The results show that the average processing time taken was 
$0.68~\unit{ms}$ and $0.74~\unit{ms}$
for the \gls{swbc} and \gls{stance} respectively. 
In both cases, the maximum computation time was always below
$2~\unit{ms}$.
}

\section{Conclusions}\label{sec:conclusion}
We presented a soft terrain adaptation algorithm called \gls{stance}:
\textbf{S}oft \textbf{T}errain \textbf{A}daptation a\textbf{N}d \textbf{C}ompliance \textbf{E}stimation.
\gls{stance} can adapt online to any type of terrain compliance (stiff or rigid). 
\gls{stance} consists of two main modules:
a compliant contact consistent whole-body controller~(\gls{awbc})
and a terrain compliance estimator~(\gls{ste}).
The \gls{awbc} 
extends our previously implemented \gls{wbc} (\gls{swbc}) \cite{Fahmi2019}, 
such that it is contact consistent
to any type of compliant terrain given the terrain parameters.
The \gls{ste} estimates online the terrain compliance and closes the loop with the \gls{awbc}. 
Unlike previous works on \gls{wbc}, \gls{stance} does not assume that the ground is rigid. 
Stance is computationally lightweight and it overcomes the limitations of the  
previous  state of the art approaches. 
As a result, \gls{stance} can efficiently traverse multiple terrains with different compliances. 
We validated \gls{stance} on our quadruped robot \gls{hyq} 
over multiple terrains of different stiffness in simulation and experiment. 
This, to the best of the authors' knowledge, is the first experimental validation on a legged robot
of closing  the loop  with a terrain estimator.

Incorporating the  terrain knowledge makes \gls{stance}  \gls{c3}.
This allows \gls{stance} to
generate smooth \grfs that are physically consistent with the terrain,
and continuously adapt the robot's feet to remain
in contact with the terrain.
As a result, 
the tracking error of the \grfs and the power consumption were reduced, 
and the impact during contact interaction was attenuated. 
Furthermore, 
\gls{stance} is more robust in challenging scenarios.
As demonstrated, \gls{stance} made it possible to perform
aggressive maneuvers and walk at high walking speeds over soft terrain
compared to the state of the art \gls{swbc}. 
In the standard case, the contact is lost because 
the motion of the terrain is not taken into account. 
On the other hand, there are minor differences
in performance between \gls{stance} and the \gls{swbc}
for less dynamic motions.

\gls{stance} can efficiently transition between multiple terrains 
with different compliances,
and each leg was able to independently 
sense and adapt to the change in terrain compliance. 
We also tested the capability of the \gls{ste} in discriminating between different terrains. 
The insights gained in simulation have been confirmed in experiment. 
 
In future works, 
we plan to implement an algorithm to improve the \gls{ste}. 
In particular, we plan on using 
onboard sensors, such as a camera,
 instead of relying on the external measurements from an MCS.
We also plan to explore other non-linear contact models in 
the \gls{ste} and the \gls{awbc}. 

%

\small
\section*{Acknowledgements}	
We thank Gustavo Medrano-Cerda, Roy Featherstone, and all the DLS lab members for the help 
provided during this work.

\bibliographystyle{./IEEEtran}
\bibliography{includes/bibliography.bib}
\vspace{-5pt}

\begin{IEEEbiography}[{\includegraphics[width=1in,height=1.28in,clip,keepaspectratio]{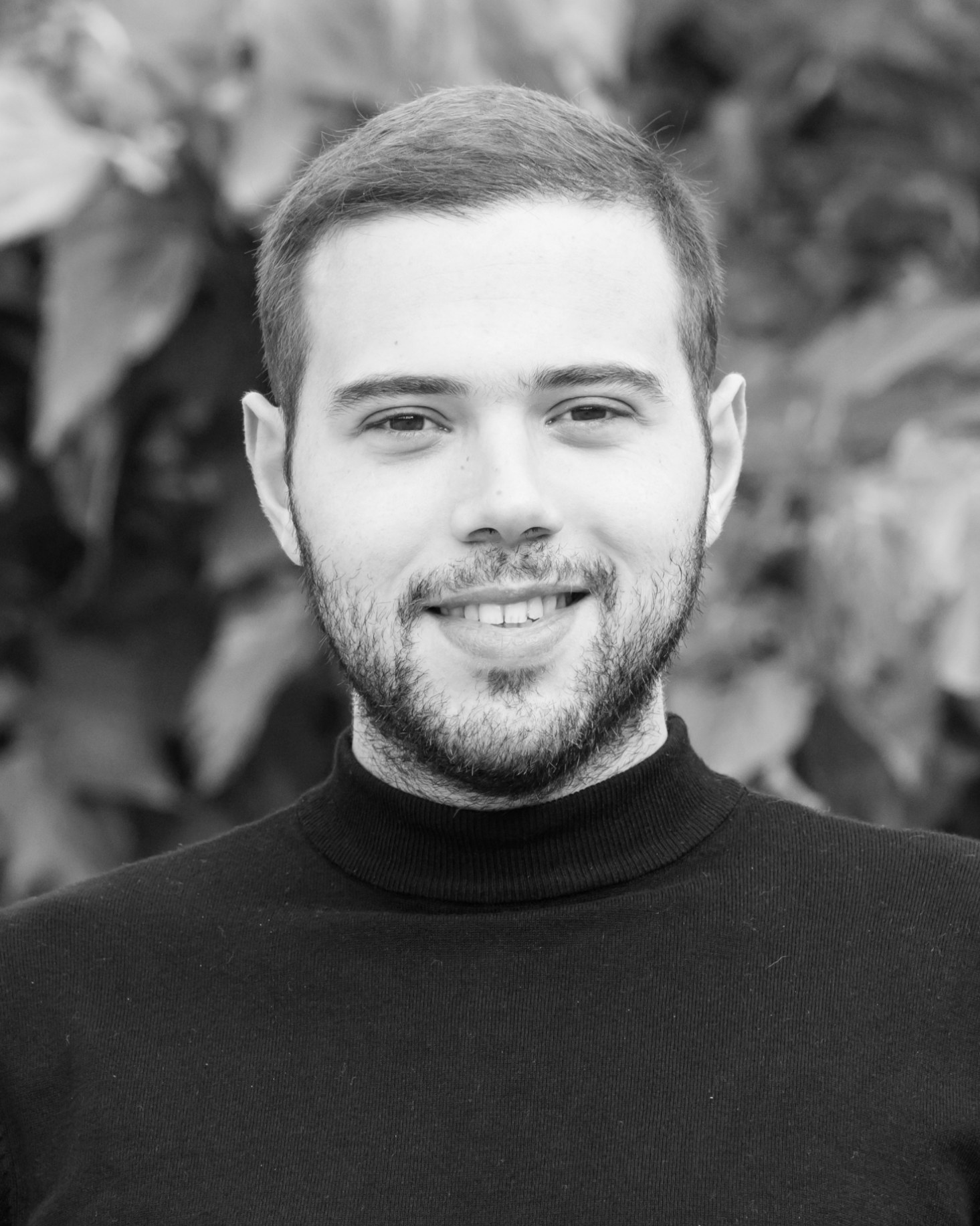}}]{Shamel Fahmi}
(S'19) was born in Cairo, Egypt. 
He received the B.Sc. degree in mechatronics from the German University in Cairo, Egypt, in 2015, and 
the M.Sc. degree in systems and control from the University of Twente, the Netherlands, in 2018. 
In December 2018, he joined the Dynamic Legged Systems (DLS) lab at Istituto Italiano di Tecnologia (IIT) for the Ph.D. degree. 
His research interests include 
robotics,
controls, 
optimization, and
learning for dynamical systems.
\end{IEEEbiography}
~
\begin{IEEEbiography}[{\includegraphics[width=1in,height=1.25in,clip,keepaspectratio]{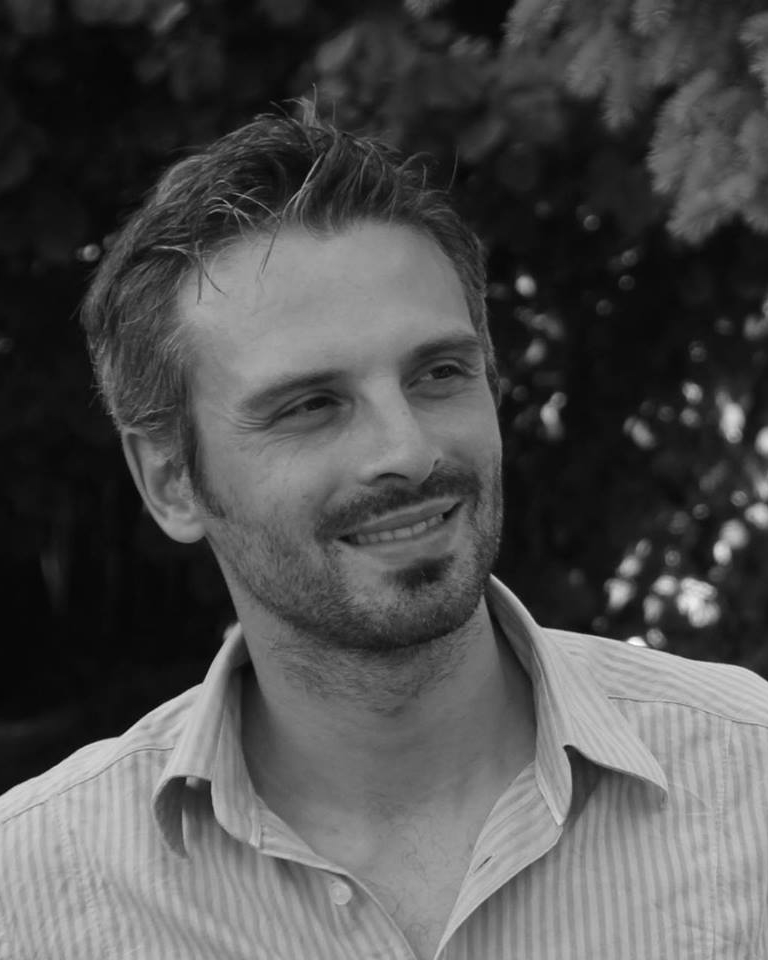}}]{Michele Focchi}
received the B.Sc. degree and the M.Sc. degree in Control System Engineering from Politecnico di Milano. 
After gaining some R$\&$D experience in the industry, in 2009, he joined Istituto Italiano di Tecnologia (IIT) 
where he developed a micro-turbine for which he obtained an international patent. 
He received the Ph.D. degree in robotics, getting involved in the Hydraulically-actuated Quadruped (HyQ) robot project, in 2013.  
He is currently a Researcher at the Dynamic Legged Systems (DLS) lab at IIT.
His research interests focus on pushing the performances of quadruped robots in traversing unstructured environments, 
by using optimization-based planning strategies to perform dynamic planning.
\end{IEEEbiography}
~
\begin{IEEEbiography}[{\includegraphics[width=1in,height=1.25in,clip,keepaspectratio]{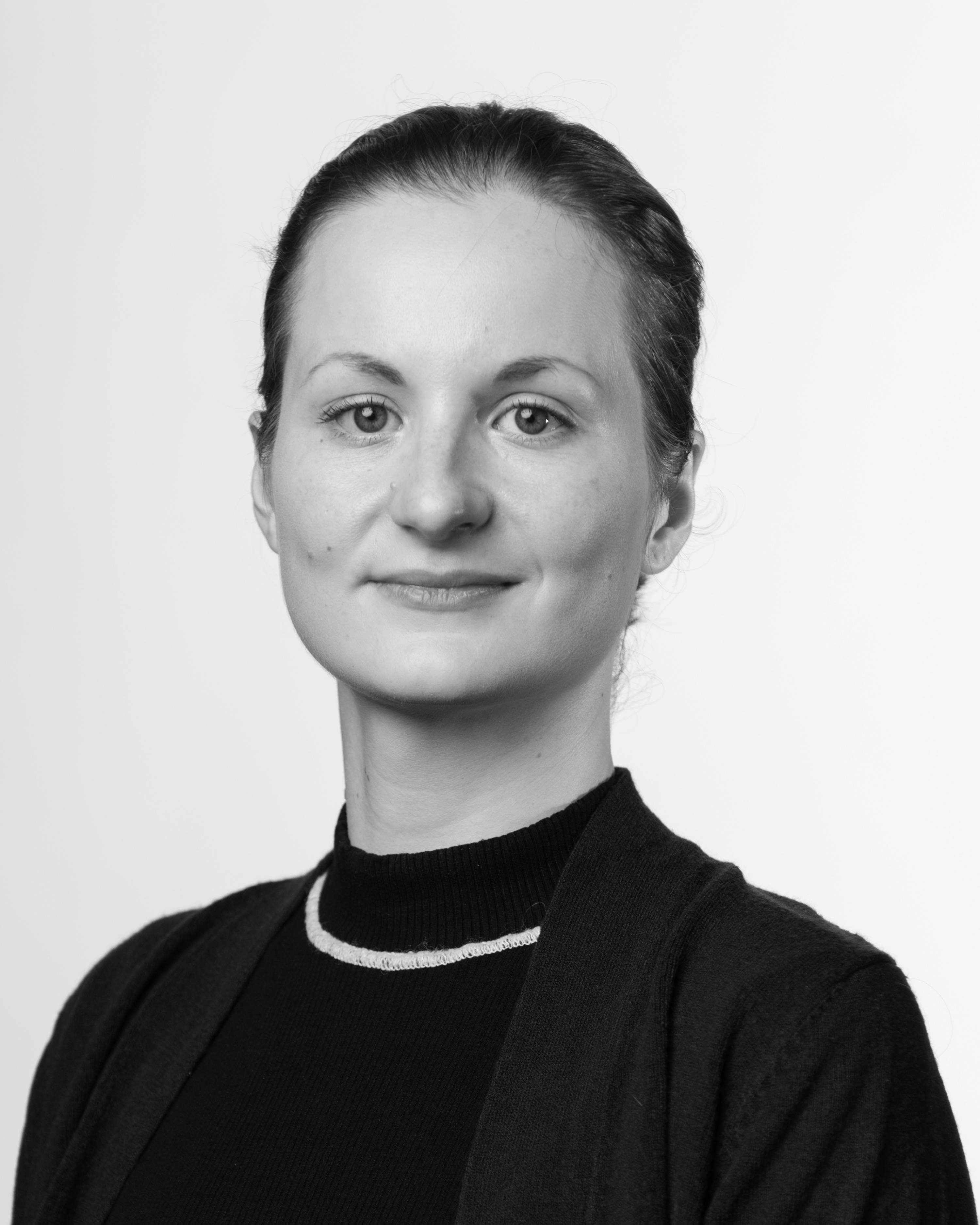}}]{Andreea Radulescu}
received the B.Sc. degree in Engineering in Automatic Control and Applied Informatics from the Polytechnic University of Bucharest, Romania. 
She received her M.Sc. degree and Ph.D degree in Intelligent Robotics from the University of Edinburgh, Scotland, in 2011 and 2016, respectively. 
From 2016 to 2019 she was a postdoc researcher in the Dynamic Legged Systems (DLS) lab at Istituto Italiano di Tecnologia (IIT). 
Her research interests include optimal control, planning, machine learning and using variable impedance actuators for systems in domains with contacts. 
She is currently a robotics research engineer at Dyson Technology Ltd.
\end{IEEEbiography}
~
\begin{IEEEbiography}[{\includegraphics[width=1in,height=1.25in,clip,keepaspectratio]{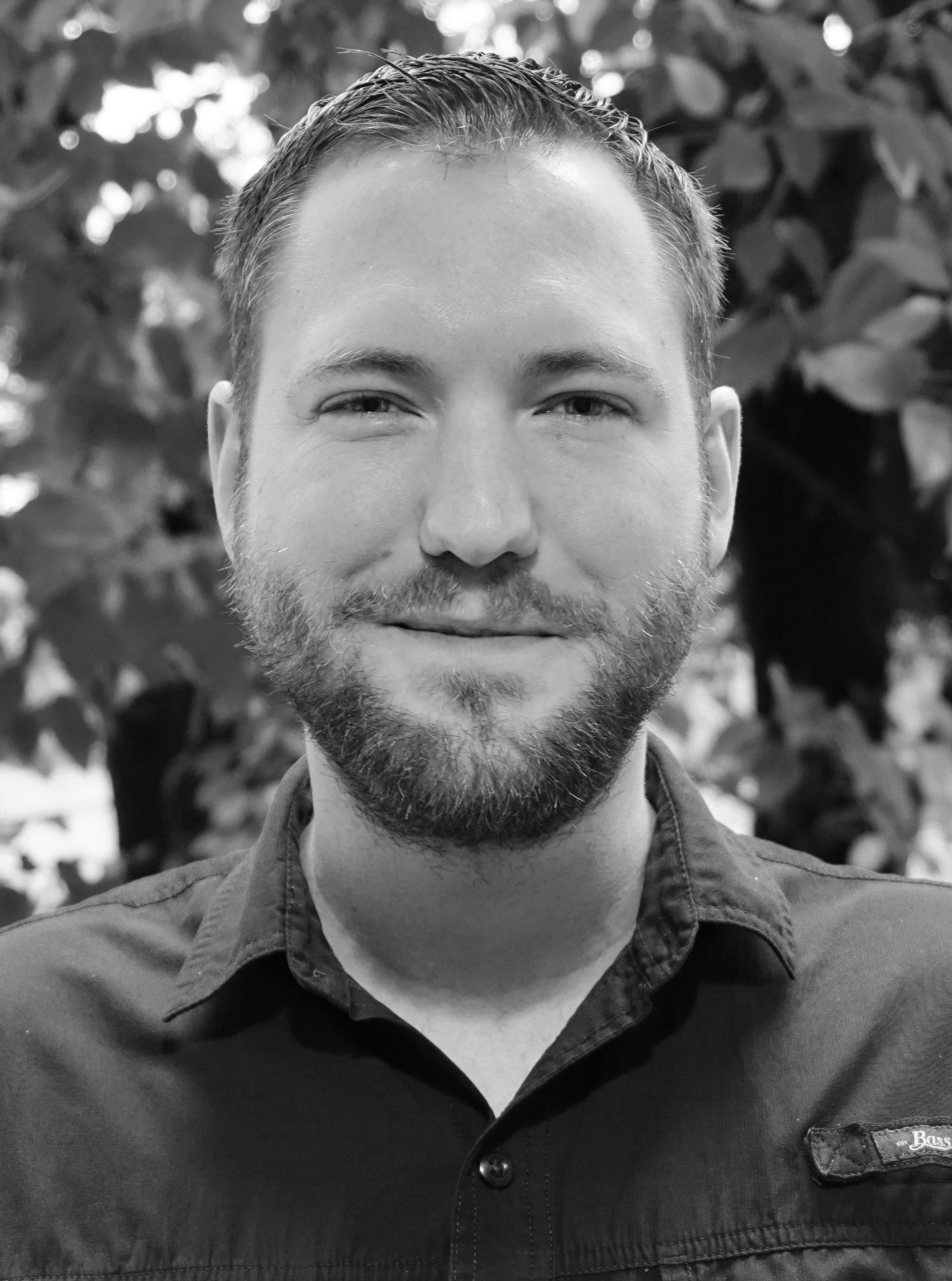}}]{Geoff Fink}
(S'08-M'18)
received the B.Sc. degree in computer engineering from the University of Alberta, Canada, in 2007, 
the M.Sc. degree in computer and electrical engineering from the University of Guadalajara, Mexico, in 2011, 
and the Ph.D. degree in control systems from the University of Alberta, Canada, in 2018. 
In 2018 he joined the the Dynamic Legged Systems (DLS) lab at Istituto Italiano di Tecnologia (IIT) as a postdoc researcher.
His research interests include robotics, sensing, perception, state estimation and SLAM.
\end{IEEEbiography}
~\newpage~
\begin{IEEEbiography}[{\includegraphics[width=1in,height=1.25in,clip,keepaspectratio]{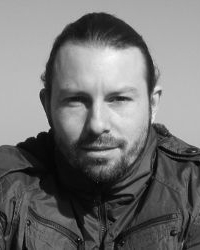}}]{Victor Barasuol}
was born in S\~{a}o Miguel do Oeste/SC, Brazil. 
He received the Diploma in electrical engineering from Universidade do Estado de Santa Catarina (UDESC) in 2006.
He received the M.Sc. degree in electrical engineering and the Ph.D degree in automation and systems engineering 
from Universidade Federal de Santa Catarina (UFSC) in 2008 and 2013, respectively. 
He is currently a researcher at the Dynamic Legged Systems (DLS) lab at Istituto Italiano di Tecnologia (IIT). 
He is an expert in motion generation and control for quadruped robots with emphasis in reactive actions using proprioceptive and exteroceptive sensor feedback. 
\end{IEEEbiography}
~
\begin{IEEEbiography}[{\includegraphics[width=1in,height=1.25in,clip,keepaspectratio]{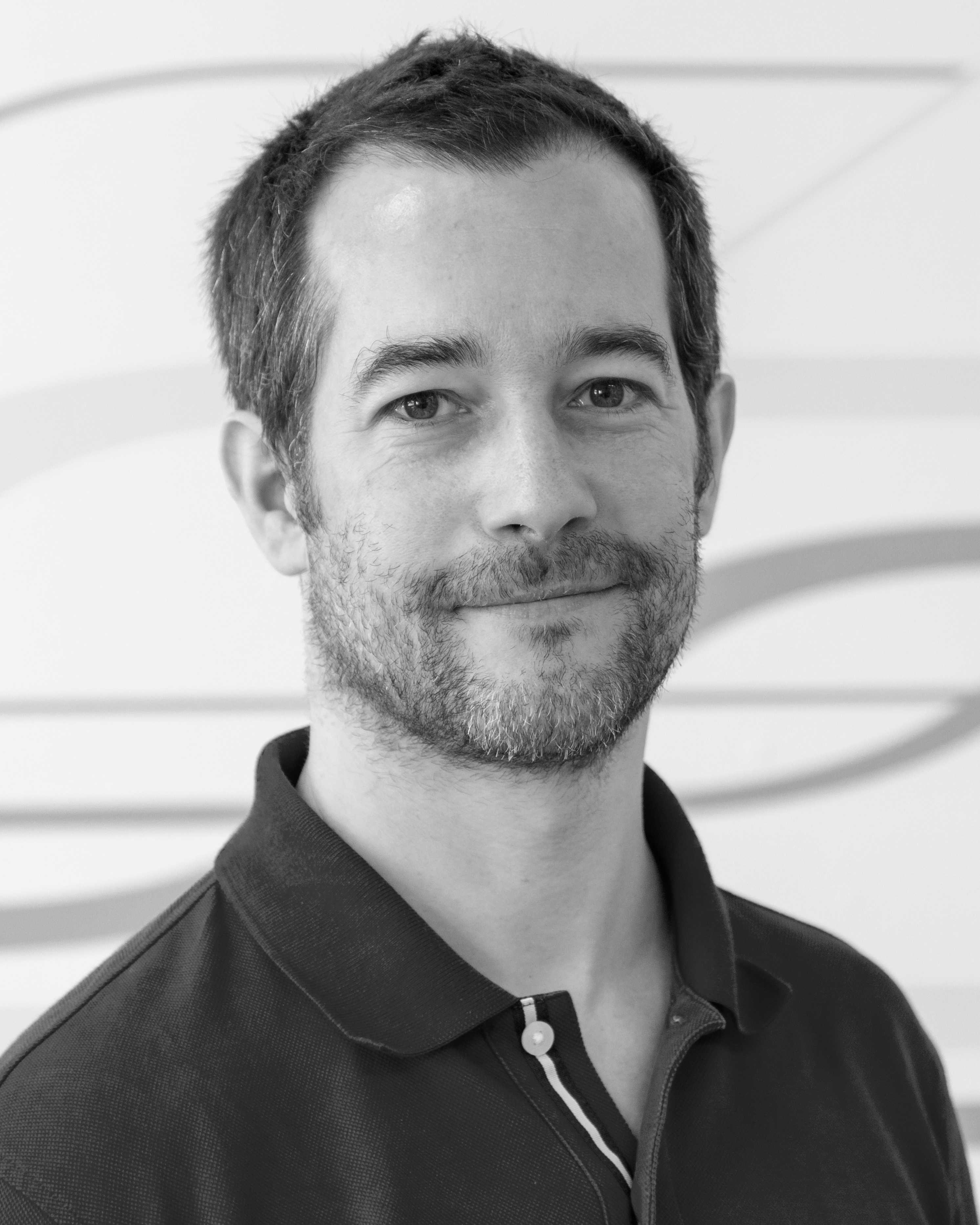}}]{Claudio Semini}
(S'07-M'10)
received the M.Sc. degree in electrical engineering and information technology from ETH, Zurich, Switzerland, in 2005, 
and the Ph.D. degree from Istituto Italiano di Tecnologia (IIT), Genoa, Italy, in 2010. 
From 2004 to 2006, he visited the Hirose Laboratory at Tokyo Tech, and worked at Toshiba’s R\&D Center, Japan. 
From 2007 to 2010, during his doctorate, he developed the hydraulic quadruped robot HyQ and worked on its control. 
From 2010 to 2012, he was a postdoc at the same department. 
He is currently a tenure-track researcher and head of the Dynamic Legged Systems (DLS) Laboratory at IIT. 
His research interests include the development and control of versatile legged robots for real-world environments.
He is a co-founder and co-chair of the IEEE RAS Technical Committee of Robot Mechanisms and Design.
\end{IEEEbiography}
\end{document}

%% file: includes/used_packages.tex
\usepackage[pdftex]{graphicx}
\usepackage[numbers]{natbib}
\usepackage{booktabs}
\usepackage{moreverb}
\usepackage{url}
\usepackage{amsmath}
\usepackage{multicol,lipsum}
\usepackage{mathtools}
\usepackage{cuted}
\usepackage{amsfonts}
\usepackage{multirow}
\usepackage{bm}
\usepackage{subfig}
\usepackage{color, colortbl}
\usepackage[colorlinks,bookmarksopen,bookmarksnumbered,citecolor=red,urlcolor=red]{hyperref}

\usepackage{enumitem}

\hypersetup
{
	pdftitle = {STANCE: Locomotion Adaptation over Soft Terrain},
	pdfauthor = {Shamel Fahmi},
	pdfsubject = {TRO manuscript},
	pdfkeywords = {Whole-body Control, Legged Robots, Compliance and Impedance Control,Optimization and 
	Optimal Control},
	pdftoolbar = true,
	colorlinks = true,
	linkcolor = black,
	citecolor = black,
	urlcolor = black,
}

\usepackage[usenames,dvipsnames]{xcolor}
\definecolor{blue_iit}{RGB}{51,51,255}
\usepackage{algpseudocode}
\usepackage{algorithm}
\usepackage{glossaries}
\usepackage[tight]{units}
\usepackage[normalem]{ulem} 
\usepackage{cancel}
\definecolor{Gray}{gray}{0.9}

\usepackage{balance}

\usepackage{flushend} 

%% file: includes/acronyms.tex
\newacronym{hyq}{HyQ}{Hydraulically actuated Quadruped}

\newacronym{lf}{LF}{Left-Front}
\newacronym{rf}{RF}{Right-Front}
\newacronym{lh}{LH}{Left-Hind}
\newacronym{rh}{RH}{Right-Hind}

\newacronym{haa}{HAA}{Hip Adduction-Abduction}
\newacronym{hfe}{HFE}{Hip Flexion-Extension}
\newacronym{kfe}{KFE}{Knee Flexion-Extension}

\newacronym{imu}{IMU}{Inertial Measurement Unit}
\newacronym{dofs}{DoFs}{Degrees of Freedom}
\newacronym{rt}{RT}{Real Time}

\newacronym{com}{CoM}{Center of Mass}
\newacronym{cop}{CoP}{Center of Pressure}
\newacronym{zmp}{ZMP}{Zero Moment Point}
\newacronym{icp}{ICP}{Instantaneous Capture Point}
\newacronym{cmp}{CMP}{Centroidal Moment Pivot}
\newacronym{grfs}{GRFs}{Ground Reaction Forces}

\newacronym{ls}{LS}{Least Square}

\newacronym{slip}{SLIP}{Spring Loaded Inverted Pendulum}
\newacronym{eom}{EoM}{Equation of Motions}
\newacronym{qp}{QP}{Quadratic Program}
\newacronym{sqp}{SQP}{Sequential Quadratic Programming}
\newacronym{mic}{MIC}{Mixed-Integer Convex}
\newacronym{cmaes}{CMA-ES}{Covariance Matrix Adaptation Evolution Strategy}
\newacronym{ara}{ARA*}{Anytime Repairing A*}
\newacronym{pca}{PCA}{Principal Component Analysis}
\newacronym{cpg}{CPG}{Central Pattern Generator}
\newacronym{wbc}{WBC}{}

\newacronym{mpc}{MPC}{Model Predictive Control}

\newacronym{awbc}{c$^3$WBC}{Compliant Contact Consistent Whole-Body Control}
\newacronym{swbc}{sWBC}{Standard Whole-Body Control}
\newacronym{c3wbc}{c$^3$WBC}{Compliant Contact Consistent Whole-Body Control}
\newacronym{ste}{TCE}{Terrain Compliance Estimator}
\newacronym{c3}{\texttt{c}$^3$}{compliant contact consistent}

\newacronym{stance}{STANCE}{}

\newacronym{wbopt}{WBOpt}{Whole-Body Optimization}

\newacronym{hc}{HC}{}
\newacronym{kv}{KV}{Kelvin-Voigt's}

\newacronym{wllsr}{WLLSR}{Weighted Linear Least Squared Regression}

\newcommand{\grfs}{\gls{grfs}~}

\newacronym{mae}{MAE}{Mean Absolute Tracking Error}

\newacronym{ode}{ODE}{Open Dynamics Engine}

%% file: includes/commands.tex
\newcommand{\Rnum}{\mathbb{R}} 

\newcommand{\grf}{F_{\mathrm{grf}}} 
\newcommand{\grfp}[1]{F_{\mathrm{grf,#1}}} 
\newcommand{\grfest}[1]{F_{\mathrm{grf},#1}} 

\newcommand{\mrm}[1]{\mathrm{#1}}
\newcommand{\nmrm}[1]{{#1}}

\newcommand{\vc}[1]{#1}
\newcommand{\mat}[1]{\ensuremath{\begin{bmatrix}#1\end{bmatrix}}}	

\newcommand{\sref}[1]{Section~\ref{#1}}
\newcommand{\eref}[1]{(\ref{#1})}
\newcommand{\fref}[1]{Fig.~\ref{#1}}
\newcommand{\tref}[1]{Table~\ref{#1}}

\newcommand\BibTeX{{\rmfamily B\kern-.05em \textsc{i\kern-.025em b}\kern-.08em
		T\kern-.1667em\lower.7ex\hbox{E}\kern-.125emX}}

\newcommand{\ie}{{i.e.},\ }
\newcommand{\eg}{{e.g.},\ }

\newcommand{\etaL}{{\textit{et~al.}}}

\newtheorem{assumpB}{Remark}

\newcommand{\assref}[1]{Algorithm~\ref{#1}}

\makeatletter
\newcounter{definition}
\newenvironment{definition}[1][t]
{\renewcommand{\ALG@name}{Formulation}
	\let\c@algocf\c@megaalgorithm
	\begin{algorithm}[#1]%
	}{\end{algorithm}}
\makeatother
\newcommand{\defref}[1]{Formulation~\ref{#1}}

\definecolor{sfahmi_blue}{RGB}{0.19,0.51,0.74}
\definecolor{LightBlue}{RGB}{0.4,0.4,1}

\newcommand{\bmcolor}[1]{\textcolor{RoyalBlue}{\bm{#1}}}

%% file: tro_sfahmi_19.bbl
\begin{thebibliography}{10}
\providecommand{\doi}[1]{DOI:\href{https://www.doi.org/#1}{#1}}
\providecommand{\url}[1]{#1}
\csname url@samestyle\endcsname
\providecommand{\newblock}{\relax}
\providecommand{\bibinfo}[2]{#2}
\providecommand{\BIBentrySTDinterwordspacing}{\spaceskip=0pt\relax}
\providecommand{\BIBentryALTinterwordstretchfactor}{4}
\providecommand{\BIBentryALTinterwordspacing}{\spaceskip=\fontdimen2\font plus
\BIBentryALTinterwordstretchfactor\fontdimen3\font minus
  \fontdimen4\font\relax}
\providecommand{\BIBforeignlanguage}[2]{{%
\expandafter\ifx\csname l@#1\endcsname\relax
\typeout{** WARNING: IEEEtran.bst: No hyphenation pattern has been}%
\typeout{** loaded for the language `#1'. Using the pattern for}%
\typeout{** the default language instead.}%
\else
\language=\csname l@#1\endcsname
\fi
#2}}
\providecommand{\BIBdecl}{\relax}
\BIBdecl

\bibitem{Farshidian2017a}
F.~{Farshidian}, E.~{Jelavić}, A.~W. {Winkler}, and J.~{Buchli}, ``Robust
  whole-body motion control of legged robots,'' in \emph{Proc. {IEEE/RSJ} Int.
  Conf. Intell. Robot. Syst. (IROS)}, Vancouver, Canada, Sep. 2017, pp.
  4589--4596,  \doi{10.1109/IROS.2017.8206328}.

\bibitem{Bellicoso2017}
C.~{Dario Bellicoso}, F.~{Jenelten}, P.~{Fankhauser}, C.~{Gehring},
  J.~{Hwangbo}, and M.~{Hutter}, ``Dynamic locomotion and whole-body control
  for quadrupedal robots,'' in \emph{Proc. {IEEE/RSJ} Int. Conf. Intell. Robot.
  Syst. (IROS)}, Vancouver, Canada, Sep. 2017, pp. 3359--3365,
  \doi{10.1109/IROS.2017.8206174}.

\bibitem{Fahmi2019}
S.~{Fahmi}, C.~{Mastalli}, M.~{Focchi}, and C.~{Semini}, ``Passive whole-body
  control for quadruped robots: Experimental validation over challenging
  terrain,'' \emph{{IEEE} Robot. Automat. Lett. ({RA-L})}, vol.~4, no.~3, pp.
  2553--2560, Jul. 2019,  \doi{10.1109/LRA.2019.2908502}.

\bibitem{Henze2017}
B.~{Henze}, A.~{Dietrich}, M.~A. {Roa}, and C.~{Ott}, ``Multi-contact balancing
  of humanoid robots in confined spaces: Utilizing knee contacts,'' in
  \emph{Proc. {IEEE/RSJ} Int. Conf. Intell. Robot. Syst. (IROS)}, Vancouver,
  Canada, Sep. 2017, pp. 697--704,  \doi{10.1109/IROS.2017.8202227}.

\bibitem{Bouyarmane2018}
K.~{Bouyarmane}, K.~{Chappellet}, J.~{Vaillant}, and A.~{Kheddar}, ``Quadratic
  programming for multirobot and task-space force control,'' \emph{{IEEE}
  Trans. Robot. ({T-RO})}, vol.~35, no.~1, pp. 64--77, Feb. 2019,
  \doi{10.1109/TRO.2018.2876782}.

\bibitem{Henze2016}
B.~Henze, M.~A. Roa, and C.~Ott, ``Passivity-based whole-body balancing for
  torque-controlled humanoid robots in multi-contact scenarios,'' \emph{Int. J.
  Robot. Res. ({IJRR})}, vol.~35, no.~12, pp. 1522--1543, Jul. 2016,
  \doi{10.1177/0278364916653815}.

\bibitem{Henze2018}
B.~{Henze}, R.~{Balachandran}, M.~A. {Roa-Garzón}, C.~{Ott}, and
  A.~{Albu-Schäffer}, ``Passivity analysis and control of humanoid robots on
  movable ground,'' \emph{{IEEE} Robot. Automat. Lett. ({RA-L})}, vol.~3,
  no.~4, pp. 3457--3464, Oct. 2018,  \doi{10.1109/LRA.2018.2853266}.

\bibitem{Azad2015}
M.~{Azad} and M.~N. {Mistry}, ``Balance control strategy for legged robots with
  compliant contacts,'' in \emph{Proc. {IEEE} Int. Conf. Robot. Automat.
  (ICRA)}, Seattle, USA, May 2015, pp. 4391--4396,
  \doi{10.1109/ICRA.2015.7139806}.

\bibitem{Vasilopoulos2018}
V.~Vasilopoulos, I.~S. Paraskevas, and E.~G. Papadopoulos, ``Monopod hopping on
  compliant terrains,'' \emph{Robot. Auton. Syst.}, vol. 102, pp. 13--26, Apr.
  2018,  \doi{10.1016/j.robot.2018.01.004}.

\bibitem{Grandia2019}
R.~{Grandia}, F.~{Farshidian}, A.~{Dosovitskiy}, R.~{Ranftl}, and M.~{Hutter},
  ``Frequency-aware model predictive control,'' \emph{{IEEE} Robot. Automat.
  Lett. ({RA-L})}, vol.~4, no.~2, pp. 1517--1524, Apr. 2019,
  \doi{10.1109/LRA.2019.2895882}.

\bibitem{Kim2019}
\BIBentryALTinterwordspacing
D.~{Kim}, S.~{Jorgensen}, J.~{Lee}, J.~{Ahn}, J.~{Luo}, and {L. Sentis},
  ``Dynamic locomotion for passive-ankle biped robots and humanoids using
  whole-body locomotion control,'' \emph{{arXiv preprint}}, pp. 1--18, 2019.
  [Online]. Available: \url{https://arxiv.org/abs/1901.08100}
\BIBentrySTDinterwordspacing

\bibitem{Neunert2018}
M.~{Neunert}, M.~{Stäuble}, M.~{Giftthaler}, C.~D. {Bellicoso}, J.~{Carius},
  C.~{Gehring}, M.~{Hutter}, and J.~{Buchli}, ``Whole-body nonlinear model
  predictive control through contacts for quadrupeds,'' \emph{{IEEE} Robot.
  Automat. Lett. ({RA-L})}, vol.~3, no.~3, pp. 1458--1465, Jul. 2018,
  \doi{10.1109/LRA.2018.2800124}.

\bibitem{Doshi2019}
N.~Doshi, K.~Jayaram, B.~Goldberg, Z.~Manchester, R.~Wood, and S.~Kuindersma,
  ``Contact-implicit optimization of locomotion trajectories for a quadrupedal
  microrobot,'' in \emph{Proc. Robot.: Sci. and Syst. (RSS)}, Pittsburgh, USA,
  2018, pp. 1--10.

\bibitem{Chang2017}
A.~H. {Chang}, C.~M. {Hubicki}, J.~J. {Aguilar}, D.~I. {Goldman}, A.~D. {Ames},
  and P.~A. {Vela}, ``Learning to jump in granular media: Unifying optimal
  control synthesis with gaussian process-based regression,'' in \emph{Proc.
  {IEEE} Int. Conf. Robot. Automat. (ICRA)}, Singapore, Singapore, May 2017,
  pp. 2154--2160,  \doi{10.1109/ICRA.2017.7989248}.

\bibitem{Alves2015}
J.~Alves, N.~Peixinho, M.~T. da~Silva, P.~Flores, and H.~M. Lankarani, ``A
  comparative study of the viscoelastic constitutive models for frictionless
  contact interfaces in solids,'' \emph{Mech. Mach. Theory}, vol.~85, pp.
  172--188, Mar. 2015,  \doi{10.1016/j.mechmachtheory.2014.11.020}.

\bibitem{Schindeler2018}
R.~{Schindeler} and K.~{Hashtrudi-Zaad}, ``Online identification of environment
  hunt–crossley models using polynomial linearization,'' \emph{{IEEE} Trans.
  Robot. ({T-RO})}, vol.~34, no.~2, pp. 447--458, Apr. 2018,
  \doi{10.1109/TRO.2017.2776318}.

\bibitem{Azad2016}
M.~{Azad}, V.~{Ortenzi}, H.~{Lin}, E.~{Rueckert}, and M.~{Mistry}, ``Model
  estimation and control of compliant contact normal force,'' in \emph{Proc.
  {IEEE}/{RAS} Int. Conf. Humanoid Robot. (Humanoids)}, Cancun, Mexico, Nov.
  2016, pp. 442--447,  \doi{10.1109/HUMANOIDS.2016.7803313}.

\bibitem{Coutinho2014}
F.~Coutinho and R.~Cortesão, ``Online stiffness estimation for robotic tasks
  with force observers,'' \emph{Control Eng. Pract.}, vol.~24, pp. 92--105,
  Mar. 2014,  \doi{10.1016/j.conengprac.2013.11.002}.

\bibitem{Coutinho2013}
F.~{Coutinho} and R.~{Cortesão}, ``A neural-based approach for stiffness
  estimation in robotic tasks,'' in \emph{Proc. Int. Conf. Adv. Robot. (ICAR)},
  Montevideo, Uruguay, Nov. 2013, pp. 1--7,  \doi{10.1109/ICAR.2013.6766499}.

\bibitem{Bosworth2016}
W.~{Bosworth}, J.~{Whitney}, {Sangbae Kim}, and N.~{Hogan}, ``Robot locomotion
  on hard and soft ground: Measuring stability and ground properties in-situ,''
  in \emph{Proc. {IEEE} Int. Conf. Robot. Automat. (ICRA)}, Stockholm, Sweden,
  May 2016, pp. 3582--3589,  \doi{10.1109/ICRA.2016.7487541}.

\bibitem{Herzog2016}
A.~Herzog, N.~Rotella, S.~Mason, F.~Grimminger, S.~Schaal, and L.~Righetti,
  ``Momentum control with hierarchical inverse dynamics on a torque-controlled
  humanoid,'' \emph{Auton. Robot.}, vol.~40, no.~3, pp. 473--491, Mar. 2016,
  \doi{10.1007/s10514-015-9476-6}.

\bibitem{Ortega2013}
R.~Ortega, J.~A.~L. Perez, P.~J. Nicklasson, and H.~J. Sira-Ramirez,
  \emph{Passivity-based control of Euler-Lagrange systems: mechanical,
  electrical and electromechanical applications}, 1st~ed., ser. Commun. Control
  Eng.\hskip 1em plus 0.5em minus 0.4em\relax Springer Science \& Business
  Media, 1998,  \doi{10.1007/978-1-4471-3603-3}.

\bibitem{Focchi2017}
M.~Focchi, A.~del Prete, I.~Havoutis, R.~Featherstone, D.~G. Caldwell, and
  C.~Semini, ``High-slope terrain locomotion for torque-controlled quadruped
  robots,'' \emph{Auton. Robot.}, vol.~41, no.~1, pp. 259--272, Jan. 2017,
  \doi{10.1007/s10514-016-9573-1}.

\bibitem{Righetti2013}
L.~Righetti, J.~Buchli, M.~Mistry, M.~Kalakrishnan, and S.~Schaal, ``Optimal
  distribution of contact forces with inverse-dynamics control,'' \emph{Int. J.
  Robot. Res. ({IJRR})}, vol.~32, no.~3, pp. 280--298, Jan. 2013,
  \doi{10.1177/0278364912469821}.

\bibitem{Tournois2017}
G.~{Tournois}, M.~{Focchi}, A.~{Del Prete}, R.~{Orsolino}, D.~G. {Caldwell},
  and C.~{Semini}, ``Online payload identification for quadruped robots,'' in
  \emph{Proc. {IEEE/RSJ} Int. Conf. Intell. Robot. Syst. (IROS)}, Vancouver,
  Canada, Sep. 2017, pp. 4889--4896,  \doi{10.1109/IROS.2017.8206367}.

\bibitem{JaeheungPark2006}
{Jaeheung Park} and O.~{Khatib}, ``Contact consistent control framework for
  humanoid robots,'' in \emph{Proc. {IEEE} Int. Conf. Robot. Automat. (ICRA)},
  Orlando, USA, May 2006, pp. 1963--1969,  \doi{10.1109/ROBOT.2006.1641993}.

\bibitem{Azad2010}
M.~Azad, R.~Featherstone \emph{et~al.}, ``Modeling the contact between a
  rolling sphere and a compliant ground plane,'' in \emph{Proc. Australas.
  Conf. Robot. Automat. (ACRA)}, Brisbane, Australia, Dec. 2010, pp. 100--107.

\bibitem{Ding2013}
L.~Ding, H.~Gao, Z.~Deng, J.~Song, Y.~Liu, G.~Liu, and K.~Iagnemma,
  ``Foot–terrain interaction mechanics for legged robots: Modeling and
  experimental validation,'' \emph{Int. J. Robot. Res. ({IJRR})}, vol.~32,
  no.~13, pp. 1585--1606, Oct. 2013,  \doi{10.1177/0278364913498122}.

\bibitem{Franklin2014}
G.~F. Franklin, J.~D. Powell, and A.~Emami-Naeini, \emph{Feedback Control of
  Dynamic Systems}, 7th~ed.\hskip 1em plus 0.5em minus 0.4em\relax Upper Saddle
  River, USA: Prentice Hall Press, 2014.

\bibitem{Stulp2015}
F.~Stulp and O.~Sigaud, ``Many regression algorithms, one unified model: A
  review,'' \emph{Neural Networks}, vol.~69, pp. 60--79, Sep. 2015,
  \doi{10.1016/j.neunet.2015.05.005}.

\bibitem{Semini2011}
C.~Semini, N.~G. Tsagarakis, E.~Guglielmino, M.~Focchi, F.~Cannella, and D.~G.
  Caldwell, ``Design of {H}y{Q} – a hydraulically and electrically actuated
  quadruped robot,'' \emph{Proc. Inst. Mech. Eng. Part I J. Syst. Control
  Eng.}, vol. 225, no.~6, pp. 831--849, 2011,  \doi{10.1177/0959651811402275}.

\bibitem{Nobili2017}
S.~Nobili, M.~Camurri, V.~Barasuol, M.~Focchi, D.~Caldwell, C.~Semini, and
  M.~Fallon, ``Heterogeneous sensor fusion for accurate state estimation of
  dynamic legged robots,'' Cambridge, USA, Jul. 2017, pp. 1--9,
  \doi{10.15607/RSS.2017.XIII.007}.

\bibitem{Focchi2018}
M.~Focchi, R.~Orsolino, M.~Camurri, V.~Barasuol, C.~Mastalli, D.~G. Caldwell,
  and C.~Semini, \emph{{Heuristic Planning for Rough Terrain Locomotion in
  Presence of External Disturbances and Variable Perception Quality}}.\hskip
  1em plus 0.5em minus 0.4em\relax Springer, 2018, vol. 132.

\bibitem{Focchi2016}
M.~Focchi, G.~A. Medrano-Cerda, T.~Boaventura, M.~Frigerio, C.~Semini,
  J.~Buchli, and D.~. Caldwell, ``Robot impedance control and passivity
  analysis with inner torque and velocity feedback loops,'' \emph{Control
  Theory Technol.}, vol.~14, no.~2, pp. 97--112, May 2016,
  \doi{10.1007/s11768-016-5015-z}.

\bibitem{Hulin2014}
T.~{Hulin}, A.~{Albu-Schäffer}, and G.~{Hirzinger}, ``Passivity and stability
  boundaries for haptic systems with time delay,'' \emph{{IEEE} Trans. Control
  Syst. Technol.}, vol.~22, no.~4, pp. 1297--1309, Jul. 2014,
  \doi{10.1109/TCST.2013.2283372}.

\bibitem{Mosadeghzad2013}
M.~Mosadeghzad, G.~A. Medrano-Cerda, N.~Tsagarakis, and D.~G. Caldwell,
  ``Impedance control with inner pi torque loop: Disturbance attenuation and
  impedance emulation,'' in \emph{Proc. Int. Conf. Robot. Biomimetics (ROBIO)},
  Shenzhen, China, Dec. 2013, pp. 1497--1502.

\bibitem{Smith2005}
\BIBentryALTinterwordspacing
R.~Smith \emph{et~al.} (2005) Open dynamics engine. [Online]. Available:
  \url{http://www.ode.org/}
\BIBentrySTDinterwordspacing

\bibitem{Catto2011}
\BIBentryALTinterwordspacing
E.~Catto, ``Soft constraints - reinventing the spring,'' in \emph{Proc. Game
  Developers Conf. (CDC)}, 2011. [Online]. Available:
  \url{http://box2d.org/files/GDC2011/GDC2011_Catto_Erin_Soft_Constraints.pdf}
\BIBentrySTDinterwordspacing

\bibitem{Erez2015}
T.~{Erez}, Y.~{Tassa}, and E.~{Todorov}, ``Simulation tools for model-based
  robotics: Comparison of {B}ullet, {H}avok, {M}u{J}o{C}o, {ODE} and
  {P}hys{X},'' in \emph{Proc. {IEEE} Int. Conf. Robot. Automat. (ICRA)},
  Seattle, USA, May 2015, pp. 4397--4404,  \doi{10.1109/ICRA.2015.7139807}.

\end{thebibliography}
